\pdfoutput=1
\documentclass[10pt]{article}

\usepackage{tmlr}

\usepackage{amsmath,amsfonts,bm}

\def\eqref#1{equation~\ref{#1}}

\def\1{\bm{1}}

\def\vc{{\bm{c}}}

\def\vh{{\bm{h}}}

\DeclareMathAlphabet{\mathsfit}{\encodingdefault}{\sfdefault}{m}{sl}
\SetMathAlphabet{\mathsfit}{bold}{\encodingdefault}{\sfdefault}{bx}{n}

\usepackage[margin=1in]{geometry}
\usepackage{placeins}
\usepackage[T1]{fontenc}
\usepackage[utf8]{inputenc}

\usepackage{amsmath}
\usepackage{amssymb}
\usepackage{amsfonts}
\usepackage{mathtools}
\usepackage{nicefrac}

\usepackage{graphicx}
\usepackage{subcaption}
\usepackage{float}

\usepackage{booktabs}
\usepackage{multirow}
\usepackage{tabularx}
\usepackage{array}

\usepackage{algorithm}
\usepackage{algpseudocode}

\usepackage{microtype}
\usepackage{xcolor}
\usepackage{comment}

\usepackage{siunitx}

\usepackage{authblk}

\usepackage{hyperref}
\usepackage{url}

\usepackage{dblfloatfix}
\title{Learning Generalizable Reconstruction of High-Dimensional Neural Dynamics}

\author{\name Anima Kujur \email anima.kujur@iwr.uni-heidelberg.de \\
      \addr Interdisciplinary Centre for Scientific Computing (IWR)\\
      Heidelberg University, Heidelberg, Germany
      \AND
      \name Zahra Monfared \email zahra.monfared@iwr.uni-heidelberg.de \\
      \addr Interdisciplinary Centre for Scientific Computing (IWR)\\
      Heidelberg University, Heidelberg, Germany\\
      Department of Mathematics and Computer Science\\
      Heidelberg University, Heidelberg, Germany}

\def\month{MM}  
\def\year{YYYY} 
\def\openreview{\url{https://openreview.net/forum?id=XXXX}} 

\begin{document}

\maketitle
\begin{abstract}
Accurate reconstruction of long-duration neural recordings is challenging because local field potentials (LFPs) combine high temporal resolution, multichannel structure, transient oscillations, and substantial variability across subjects. Conventional Dynamic Mode Decomposition (DMD) methods often become poorly conditioned or lose fine-scale temporal structure on long, high-dimensional signals, whereas deep learning approaches may require substantial computational resources and provide limited dynamical interpretability. We present PCA-DMD, a scalable operator-theoretic framework that segments multichannel LFP recordings into overlapping windows, projects them into a compact principal-component space, learns a linear approximation of Koopman evolution in the latent space, and reconstructs the continuous signal through inverse projection and overlap-add aggregation. We evaluate the framework through progressively more demanding experiments. On $200{,}000$-sample hippocampal recordings, PCA-DMD outperformed Classical DMD, SpDMD, MrDMD, and HODMD, achieving a Kullback--Leibler divergence (KLD) of $0.0761$ and a Hellinger distance (HD) of $0.0847$. In all-pair cross-subject zero-shot generalization at $300{,}000$ samples, PCA-DMD achieved correlations of $0.9504$--$0.9800$, with HD values of $0.0010$--$0.0072$ and KLD values of $0.0005$--$0.0022$, without target-subject fine-tuning. The prediction further showed that the learned latent dynamics accurately predict temporally held-out LFP segments excluded from model fitting, maintaining close one-step prediction agreement across the unseen interval, multiple channels, and temporally separated early, middle, and late regions. A dedicated scalability analysis, in which models fitted on S1 were transferred to S2--S4, showed stable zero-shot reconstruction from $400{,}000$ to $900{,}000$ samples, with mean correlation remaining approximately $0.965$--$0.968$ while computational cost increased predictably. External validation on an independent 93-channel Allen Neuropixels recording yielded a mean channel-wise correlation of $0.7427$ and a median correlation of $0.7990$. Koopman spectral and mode analyses further revealed stable latent dynamics with dominant eigenvalues concentrated near the unit circle. These results establish PCA-DMD as an interpretable, generalizable, and computationally scalable framework for reconstruction of high-dimensional neural dynamics, demonstrating cross-subject zero-shot generalization together with accurate prediction.
\end{abstract}

\section{Introduction}
\label{sec:introduction}

High-dimensional time series arise throughout science, including neuroscience, biology, climate science, and engineering, where a central objective is to discover compact representations that preserve the essential structure of the underlying dynamics. This challenge is particularly pronounced in neural systems because modern electrophysiological recordings combine high temporal resolution, multiple simultaneously recorded channels, transient activity, nonlinear interactions, and long observation horizons. Accurate reconstruction of such recordings is therefore more than a signal-recovery problem: it provides a demanding test of whether a model has captured meaningful dynamical structure that remains valid over extended periods and across changing experimental conditions.

Hippocampal local field potentials (LFPs) provide a particularly challenging example of this class of data. LFPs reflect the extracellular activity generated by neuronal populations and provide a mesoscopic representation of circuit-level dynamics \cite{molle2006hippocampal,buzsaki2012origin,einevoll2013modelling,meier2020LFPMice,safavi2023uncovering,cimbalnik2025LFPRecording}. These recordings exhibit coordinated activity across multiple channels and temporal scales \cite{Buzsaki2012,Buzsaki2015}. Of particular interest are hippocampal sharp wave-ripples (SWRs), brief high-frequency oscillatory events in the approximate range of $100$--$250$ Hz that occur together with sharp waves lasting approximately $50$--$100$ ms. SWRs are associated with memory consolidation, neural replay, planning, and information transfer between brain regions \cite{joo2018hippocampal,jin2024sWR,liao2024Hippocampal,buzsaki2015SWR,leonard2015sharpHippo}. They arise from highly coordinated neuronal activity and exhibit complex multiscale organization across space and time \cite{buzsaki2010neural,logothetis2012hippocampal,kaplan2016hippocampal,abadchi2020spatiotemporal,safavi2021univariate,nitzan2022brain}.

Reliable reconstruction of LFP recordings containing SWRs is important for both computational neuroscience and downstream signal analysis. Reconstruction can recover temporal structure that may be obscured by noise, artifacts, or incomplete observations while preserving transient oscillations that are easily lost through aggressive smoothing or downsampling. Reconstructed trajectories can also support the extraction of dynamical quantities such as Koopman eigenvalues, modes, eigenfunctions, oscillatory frequencies, and stability characteristics \cite{wasim2025}. Such representations may provide a more mechanistic description of neural activity than reconstruction accuracy alone and may ultimately support applications involving neural-state characterization and the study of pathological ripple activity in disorders such as epilepsy, schizophrenia, and Alzheimer's disease \cite{buzsaki2015SWR}.

Despite the biological importance of SWRs, most existing studies have focused on their detection, classification, or prediction rather than on reconstruction of the underlying long-duration multichannel signals. Representative approaches include convolutional neural networks for automated SWR detection \cite{navas2022SWVRPrediction}, recurrent architectures such as RippleNet \cite{hagen2021ripplenetswrdetection}, consensus frameworks for distinguishing physiological SWRs from other high-frequency oscillations \cite{liu2022SWRFastOscillation}, and spatiotemporal models for separating SWRs from epileptiform discharges \cite{maslarova2025spatiotemporalSWRMain}. Additional studies have investigated topological properties of SWR waveforms \cite{sebastian2023topologicalSWRimp}, machine-learning toolboxes for cross-species analysis \cite{navas2024swrMLAnalysis}, singular-value-based artifact removal \cite{chen2025svdLFP}, and reconstruction methods for related electrophysiological signals such as EEG and MEG \cite{cao2022eegReconstruction}. However, direct reconstruction of long-horizon hippocampal LFP signals while retaining transient SWR-related structure remains comparatively underexplored.

The scale of modern neural recordings makes this reconstruction problem especially difficult. A single multichannel LFP recording may contain millions of samples, while the underlying dynamics are transient, nonstationary, and heterogeneous across subjects. Deep recurrent architectures, including long short-term memory networks \cite{hochreiter1997long}, were developed in part to alleviate the vanishing- and exploding-gradient problems of conventional recurrent neural networks \cite{bengio1994learning}. Nevertheless, training over extremely long sequences can remain difficult because gradient propagation becomes increasingly ill-conditioned, memory consumption grows with sequence length, and small parameter changes may produce substantial variations in long-horizon output \cite{pascanu2013difficulty}. Structured latent recurrent models, including shPLRNN-based formulations, can capture nonlinear neural dynamics but remain sensitive to latent dimensionality, teacher forcing, and other hyperparameters when extended to long recordings \cite{hess_generalized_2023}.

Deep models also depend strongly on the availability and representativeness of training data. In limited-data or distribution-shift settings, highly expressive networks can fit spurious regularities that do not correspond to the underlying physiological dynamics \cite{zhang2021understanding}. This issue is particularly relevant for neural recordings, where signal statistics vary across animals, recording sessions, electrodes, and acquisition systems. Consequently, a representation learned from one subject may not necessarily generalize to an unseen subject without adaptation. Moreover, the internal representations of deep networks generally do not map directly to interpretable dynamical quantities such as oscillatory frequencies, growth or decay rates, and stability modes \cite{lipton2018mythos}. These limitations do not preclude the use of deep learning for neural signals, but they motivate complementary approaches that are computationally efficient, capable of cross-subject generalization, and explicitly connected to dynamical-systems theory.

Koopman operator theory provides such a framework by representing nonlinear dynamical systems through linear evolution in a lifted space of observable functions \cite{Koopman1931}. Dynamic Mode Decomposition (DMD) offers a practical, data-driven finite-dimensional approximation of Koopman spectral analysis and has been widely used to identify coherent spatiotemporal structures in complex systems \cite{Schmid2010,Rowley2009,Kutz2016}. Its eigenvalues and modes provide direct information about oscillation, persistence, growth, decay, and spatial organization. However, direct application of DMD to long, high-dimensional neural recordings remains challenging. Sliding-window representations of multichannel LFPs produce large data matrices that may be poorly conditioned, leading to unstable regression and unreliable spectral estimates \cite{tu2014dynamic}. Rank truncation can reduce computational cost but may discard high-frequency components that are essential for representing transient SWRs.

Conventional batch DMD methods also scale poorly as the recording horizon and observation dimension increase. Streaming and low-storage formulations have been developed for large or continuously arriving datasets \cite{hemati2014dynamic}, but general-purpose DMD implementations can still exhibit substantial memory demands, loss of reconstruction fidelity, and numerical instability on very long neural recordings. Furthermore, conventional DMD models are generally fitted independently to each recording and do not explicitly address whether a learned dynamical representation generalizes to unseen subjects. Cross-subject generalization is particularly challenging because neural recordings are subject to substantial inter-subject and inter-session distribution shifts \cite{farshchian2019adversarial}. These limitations motivate a framework that retains the interpretability of Koopman analysis while improving numerical conditioning, scalability, and generalization across subjects.

To address these challenges, PCA-DMD combines principal component analysis (PCA) with Koopman-inspired latent evolution \cite{Kujur2025}. Multichannel LFP recordings are first segmented into overlapping temporal windows and vectorized into high-dimensional observations. PCA projects these observations into a compact latent subspace, where a linear operator is estimated to describe the evolution between consecutive windows. Predicted latent states are subsequently mapped back to the original observation space and aggregated through overlap-add reconstruction. The PCA stage reduces the dimensionality of the Koopman regression, suppresses redundant directions, and improves numerical conditioning, while the overlap-add procedure preserves continuity across neighboring windows.

Although PCA-DMD is related to Hankel-DMD, delay-coordinate DMD, and HAVOK-type formulations, its computational organization is distinct. PCA is applied as an explicit pre-compression step to the vectorized multichannel windows; Koopman evolution is then learned directly in the resulting PCA latent space; and the complete recording is reconstructed through inverse projection and overlap-add aggregation. This separation between compression, latent evolution, and signal assembly provides a practical mechanism for modeling long neural recordings while retaining interpretable spectral structure.

The present study provides a unified and substantially expanded evaluation of PCA-DMD for long-horizon neural signal reconstruction. The evaluation comprises five complementary experimental settings. First, reconstruction fidelity is assessed on $200{,}000$-sample hippocampal recordings through comparison with Classical DMD, Sparse DMD (SpDMD), Multi-resolution DMD (MrDMD), Higher-Order DMD (HODMD), and Hankel-DMD. Second, cross-subject generalization is examined through an all-pair zero-shot evaluation at $300{,}000$ samples, in which models fitted on one subject are applied directly to unseen target subjects without target-specific fine-tuning. Third, the prediction is evaluated on temporally held-out LFP segments excluded from PCA and DMD fitting to determine whether the learned latent evolution predicts unseen neural states. Fourth, a dedicated scalability analysis examines S1-to-S2/S3/S4 zero-shot transfer as signal length increases from $400{,}000$ to $900{,}000$ samples. Fifth, external validation is performed on an independent 93-channel Allen Neuropixels LFP recording with different dimensionality, sampling rate, and acquisition characteristics.

The results demonstrate complementary strengths across these settings. At $200{,}000$ samples, PCA-DMD achieves the lowest Hellinger distance (HD) and Kullback--Leibler divergence (KLD) among the evaluated DMD variants. In the $300{,}000$-sample all-pair zero-shot experiment, correlations range from $0.9504$ to $0.9800$, with low distributional discrepancies across source--target pairs, demonstrating strong cross-subject generalization without target-specific fine-tuning. The prediction further shows close one-step agreement on temporally held-out LFP segments, across the full unseen interval, multiple channels, and temporally separated early, middle, and late regions. The scalability experiments show that mean correlation remains approximately $0.965$--$0.968$ as sequence length increases from $400{,}000$ to $900{,}000$ samples, while runtime grows predictably. On the independent Allen recording, PCA-DMD achieves a mean channel-wise correlation of $0.7427$ and a median correlation of $0.7990$ across 93 channels. Koopman spectral and mode analyses further reveal dominant eigenvalues concentrated near the unit circle and consistent latent organization across subjects.

The main contributions of this work are summarized as follows:

\begin{itemize}

    \item A scalable PCA-DMD framework for reconstruction of long-duration multichannel neural recordings, combining PCA-based compression, latent Koopman evolution, inverse projection, and overlap-add aggregation while preserving transient spatiotemporal LFP structure.

    \item A systematic $200{,}000$-sample reconstruction benchmark against Classical DMD, SpDMD, MrDMD, HODMD, and Hankel-DMD, demonstrating improved signal fidelity and distributional agreement.

    \item An all-pair cross-subject zero-shot generalization protocol across four subjects at $300{,}000$ samples, demonstrating that the learned PCA representation and latent Koopman dynamics can be transferred to unseen subjects without target-specific fine-tuning.

    \item The prediction on held-out LFP segments excluded from PCA and DMD fitting, assessing one-step prediction of unseen neural states across the full temporal interval, multiple channels, and temporally separated early, middle, and late regions.

    \item A long-horizon scalability analysis from $400{,}000$ to $900{,}000$ samples using S1 as the source and S2--S4 as unseen targets, jointly evaluating reconstruction accuracy, distributional agreement, normalized reconstruction error, and computational cost.

    \item External validation on an independent 93-channel Allen Neuropixels recording, demonstrating applicability of the reconstruction framework under substantially different channel dimensionality, sampling rate, and acquisition characteristics.

    \item Koopman spectral and mode-level analyses connecting reconstruction behavior with the stability, persistence, and organization of the learned latent dynamics.


\end{itemize}

\section{Related Work}
\label{sec:related_work}

\subsection{Hippocampal LFPs and Sharp Wave-Ripples}
\label{sec:related_lfp}

Local field potentials (LFPs) provide a mesoscopic description of coordinated neural population activity and are widely used to study circuit-level computation \cite{Buzsaki2012,Buzsaki2015,buzsaki2012origin}. Hippocampal LFPs exhibit structured oscillatory activity across multiple spatial and temporal scales, including theta, gamma, and sharp wave-ripple regimes associated with navigation, memory, and neural communication \cite{Joo2018,Wilson1994,Diba2007,Foster2017,colgin2016rhythms}. During sharp wave-ripples (SWRs), fast oscillatory activity is accompanied by highly synchronized neuronal firing, particularly within hippocampal CA1 networks \cite{csicsvari1999fast}.

SWRs have been linked to memory consolidation, replay, retrieval, and planning \cite{joo2018hippocampal,buzsaki2015SWR,leonard2015sharpHippo}. Their functional relevance is supported by intervention studies showing that selective suppression of ripple activity during post-learning periods impairs hippocampus-dependent memory performance \cite{girardeau2009selective}. These findings make SWRs an important target for computational modeling, but their brief duration, high-frequency content, and variability across channels make faithful long-duration reconstruction difficult.

Most computational studies have concentrated on detecting or classifying individual SWR events rather than reconstructing the continuous multichannel signals in which they occur. Existing methods include convolutional and recurrent models for automated detection \cite{navas2022SWVRPrediction,hagen2021ripplenetswrdetection}, consensus and spatiotemporal frameworks for distinguishing physiological ripples from pathological high-frequency oscillations \cite{liu2022SWRFastOscillation,maslarova2025spatiotemporalSWRMain}, and topological or cross-species tools for waveform characterization \cite{sebastian2023topologicalSWRimp,navas2024swrMLAnalysis}. Although these approaches have advanced event-level analysis, scalable reconstruction of the underlying long-horizon hippocampal LFP remains comparatively underexplored.

\subsection{Neural Time-Series Reconstruction and Latent Dynamical Models}
\label{sec:related_latent_models}

A broad class of neural time-series methods seeks low-dimensional latent representations that explain high-dimensional population activity \cite{Cunningham2014,Pandarinath2018,Durstewitz2023}. These methods include probabilistic state-space models, recurrent neural networks, variational dynamical systems, and structured latent recurrent architectures.

Latent Factor Analysis via Dynamical Systems (LFADS), for example, uses recurrent networks to infer smooth latent trajectories from single-trial population recordings \cite{sussillo2016lfads}. Recurrent switching linear dynamical systems combine continuous latent evolution with discrete regime transitions, providing piecewise-linear descriptions of complex neural dynamics \cite{linderman2017bayesian}. Structured recurrent models such as generalized piecewise-linear recurrent neural networks further attempt to balance nonlinear expressiveness with dynamical interpretability \cite{hess_generalized_2023}.

These approaches can recover informative latent trajectories, but their application to very long neural recordings remains challenging. Recurrent optimization becomes increasingly difficult as sequence length grows, and memory requirements can limit the use of full-resolution recordings \cite{pascanu2013difficulty,pmlr-v162-brenner22a,Brenner2024}. Truncation, downsampling, and segmented training can reduce the computational burden, but they may weaken long-range temporal consistency or remove transient high-frequency events. Structured latent models can also be sensitive to latent dimension, regularization, and teacher-forcing settings \cite{hess_generalized_2023}. These limitations motivate non-recurrent formulations that retain compact latent representations while supporting direct and interpretable dynamical evolution.

\subsection{Koopman Operator Theory and Dynamic Mode Decomposition}
\label{sec:related_koopman}

Koopman operator theory represents nonlinear dynamical systems through linear evolution in a generally infinite-dimensional space of observable functions \cite{Koopman1931}. This formulation permits nonlinear dynamics to be analyzed using spectral quantities such as eigenvalues, eigenfunctions, and modes. Dynamic Mode Decomposition (DMD) provides a finite-dimensional, data-driven approximation of this spectral representation and has been widely applied to extract coherent spatiotemporal structures from complex systems \cite{Schmid2010,Rowley2009,Kutz2016}. The interpretation of DMD eigenvalues and modes in terms of oscillation, growth, decay, and persistence is a central advantage of the approach \cite{Brunton2022,Schmid2022}.

The standard DMD algorithm estimates a best-fit linear evolution operator from paired observations \cite{tu2014dynamic}. Several extensions have been developed to enrich the observable space or improve the resulting representation. Extended DMD introduces a dictionary of nonlinear observables to approximate a broader class of Koopman eigenfunctions \cite{williams2015data}. Delay-coordinate and Hankel formulations augment the state with temporally shifted observations, with theoretical convergence properties established for ergodic systems \cite{arbabi2017ergodic}. HAVOK similarly uses delay embeddings to separate approximately linear evolution from intermittent forcing \cite{Brunton2017}.

Other established variants address particular structural or computational objectives. Sparsity-Promoting DMD (SpDMD) selects a parsimonious subset of dynamically relevant modes \cite{Jovanovic2014}. Higher-Order DMD (HODMD) uses multiple temporal lags to improve the representation of complex and multi-frequency dynamics \cite{LeClainche2017}. Multi-Resolution DMD (MrDMD) recursively separates activity across temporal scales \cite{Kutz2016MrDMD}. Streaming and low-storage formulations update the decomposition incrementally and reduce the need to retain the complete observation history \cite{hemati2014dynamic}.

Despite this extensive methodological development, direct DMD analysis of long multichannel neural recordings remains difficult. Vectorized sliding windows produce high-dimensional and potentially ill-conditioned regression problems, while rank truncation may remove short-lived or high-frequency signal components. Moreover, most DMD studies fit an operator independently to each recording; comparatively little attention has been given to whether a dynamical representation learned from one subject generalizes to unseen subjects without adaptation.

\subsection{Scalable and Cross-Subject Generalizable Neural Representations}
\label{sec:related_transfer}

Dimensionality reduction is central to neural data analysis because population recordings often occupy a substantially lower-dimensional space than their observation dimension suggests \cite{Cunningham2014}. Principal component analysis (PCA) remains widely used because it is computationally efficient, interpretable, and capable of retaining dominant variance directions. When dimensionality reduction precedes dynamical regression, it can reduce memory requirements, suppress redundant directions, and improve numerical conditioning.

A related challenge is whether a learned neural representation generalizes across subjects, recording sessions, or acquisition conditions. Neural signals commonly exhibit distribution shifts caused by anatomical variability, electrode placement, measurement noise, and changes in physiological state. In brain--computer interface applications, adversarial domain-adaptation methods have been used to align representations across recording sessions \cite{farshchian2019adversarial}. Similar variability has motivated cross-subject and cross-task representation learning for electroencephalography (EEG) \cite{eegcrosssubjectsurvey2026}. Recent zero-shot EEG studies further indicate that models with strong within-domain performance can generalize poorly when applied directly to unseen subjects \cite{zeroshoteeg2026}.

Most approaches to cross-subject neural generalization rely on deep domain-invariant representations. In contrast, cross-subject generalization of data-driven Koopman representations has received substantially less attention. Existing work on DMD scalability has primarily addressed storage and incremental computation within a single data stream \cite{hemati2014dynamic}, rather than determining whether an estimated latent dynamical operator remains effective when applied directly to unseen subjects.

PCA-DMD was introduced as a scalable Koopman-based framework for reconstructing long hippocampal LFP recordings \cite{Kujur2025,kujur2026learning}. It projects vectorized temporal windows into a compact PCA space, estimates linear evolution in that latent space, and reconstructs the full signal through inverse projection and overlap-add aggregation. Initial experiments demonstrated improved reconstruction of recordings containing up to $200{,}000$ samples relative to conventional DMD variants.

The present work extends PCA-DMD beyond the original within-subject reconstruction setting along several complementary dimensions. Cross-subject generalization is evaluated through all ordered source--target pairs at $300{,}000$ samples, with the model learned from each source subject applied directly to unseen target subjects without target-specific fine-tuning. The prediction is separately examined on temporally held-out LFP segments excluded from model fitting. Scalability is assessed through S1-to-S2/S3/S4 zero-shot transfer as signal length increases from $400{,}000$ to $900{,}000$ samples, while robustness beyond the primary dataset is evaluated on an independent 93-channel Allen Neuropixels recording. Together, these experiments examine reconstruction fidelity, cross-subject generalization, prediction, sequence-length scalability, and external-dataset robustness within a unified operator-theoretic framework.

\section{Materials and Data}
\label{sec:data}

\subsection{Primary Hippocampal LFP Dataset}
\label{sec:primary_data}

We evaluate the proposed framework using hippocampal local field potential (LFP) recordings from rodents collected for sharp wave-ripple (SWR) analysis \cite{meier2020LFPMice}. The recordings were obtained from the publicly available repository introduced in \cite{navas2024machineData}, which provides multichannel hippocampal signals together with expert annotations of SWR events.

The dataset contains recordings from four subjects, denoted by
\[
\mathcal{S}=\{S1,S2,S3,S4\}.
\]
Each subject contains \(C=8\) simultaneously recorded LFP channels sampled at
\[
f_s=30{,}000~\mathrm{Hz}.
\]
This sampling rate is sufficient to resolve the high-frequency ripple activity typically observed in the \(100\)--\(250~\mathrm{Hz}\) range. Each channel contains approximately \(22{,}326{,}272\) samples, corresponding to roughly \(744\) seconds of continuous recording.

The primary dataset is used in four complementary experimental settings. First, within-subject reconstruction is evaluated on \(200{,}000\)-sample segments to compare PCA-DMD with conventional DMD variants. Second, cross-subject zero-shot generalization is evaluated on \(300{,}000\)-sample segments using all ordered source--target subject pairs, with no target-specific fine-tuning. Third, the prediction is evaluated on temporally held-out LFP segments excluded from PCA and DMD fitting, providing a separate assessment of one-step prediction on unseen temporal data. Fourth, a dedicated scalability analysis is performed using signal lengths of
\[
400{,}000,\ 500{,}000,\ 600{,}000,\ 700{,}000,\ 800{,}000,\ \text{and}\ 900{,}000
\]
samples. In this setting, PCA-DMD is fitted on source subject \(S1\) and applied without fine-tuning to the unseen target subjects \(S2\), \(S3\), and \(S4\).

Together, these experimental settings assess reconstruction fidelity, cross-subject generalization, prediction, and scalability while keeping the underlying acquisition setting fixed.

\subsection{External Allen Neuropixels Validation Dataset}
\label{sec:allen_data}

To assess robustness beyond the primary hippocampal dataset, we additionally evaluate PCA-DMD on an independent Allen Neuropixels LFP recording from the Allen Brain Observatory Visual Coding dataset \cite{allen_visual_behavior_neuropixels_2024,bennett2025visualbehaviorneuropixels}. The external recording contains \(93\) channels and approximately \(300{,}000\) samples acquired at a sampling rate of approximately \(1{,}250~\mathrm{Hz}\), corresponding to roughly \(240\) seconds of neural activity.

The Allen recording differs substantially from the primary dataset in channel dimensionality, sampling rate, recording duration, and acquisition setting. It is therefore used for external validation to determine whether the same PCA-DMD reconstruction mechanism remains effective under a distinct neural recording configuration. The external experiment evaluates full-length reconstruction, channel-wise performance, multichannel structure, PCA explained variance, and Koopman spectral properties.

\section{Methods}
\label{sec:methods}

\begin{figure*}[!t]
    \centering
    \includegraphics[width=\textwidth]{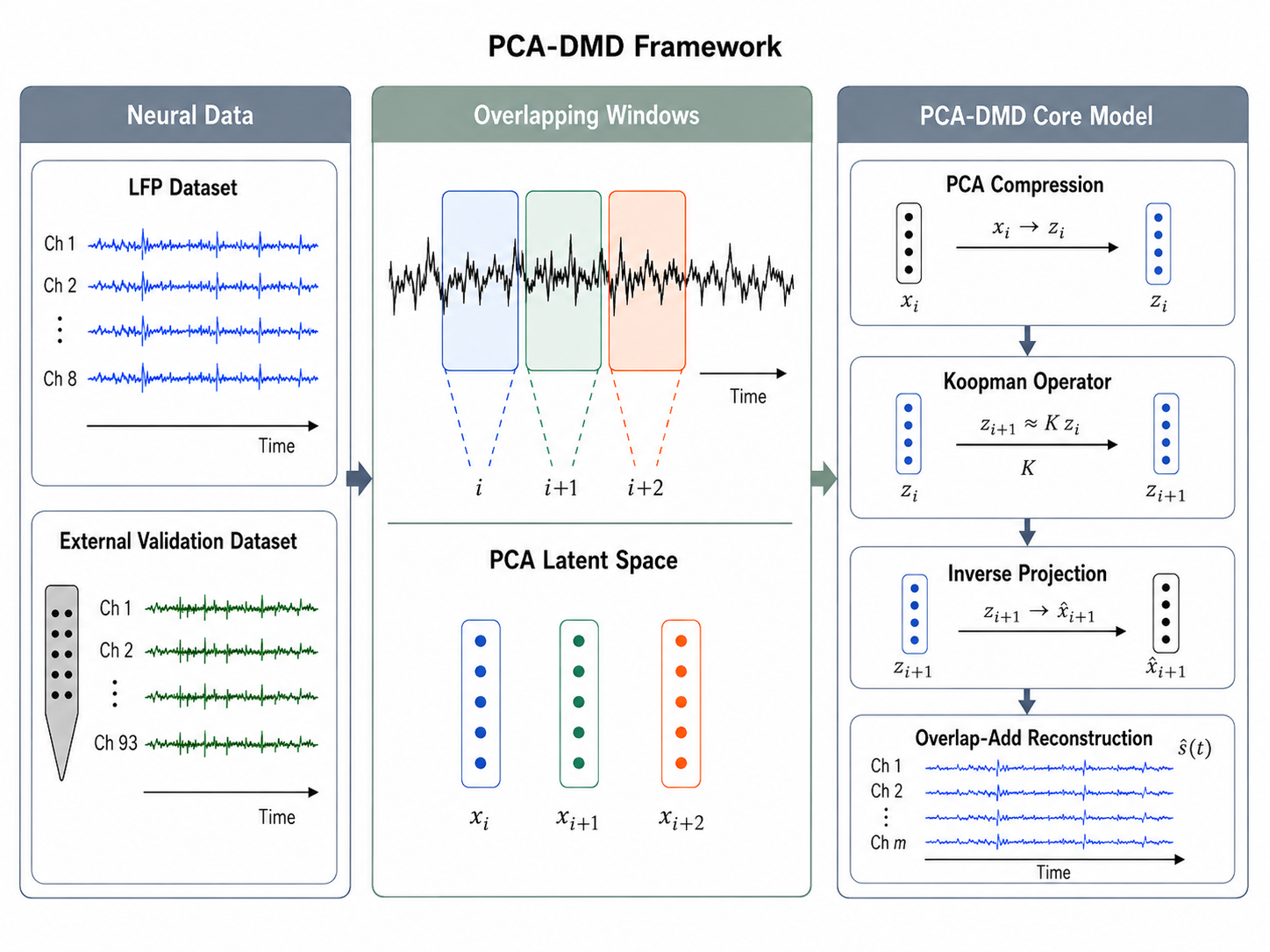}
    \caption{
    Overview of the PCA-DMD computational framework. Multichannel LFP
    recordings are segmented into overlapping temporal windows and
    vectorized into high-dimensional snapshot representations. PCA projects
    the snapshots into a compact latent space, where a linear Koopman/DMD
    operator models evolution between consecutive latent states. Predicted
    latent states are mapped back to the original signal space through
    inverse PCA projection, and overlapping reconstructed windows are
    aggregated to recover a continuous multichannel signal.
    }
    \label{fig:framework}
\end{figure*}

\begin{figure*}[!t]
    \centering
    \includegraphics[width=\textwidth]{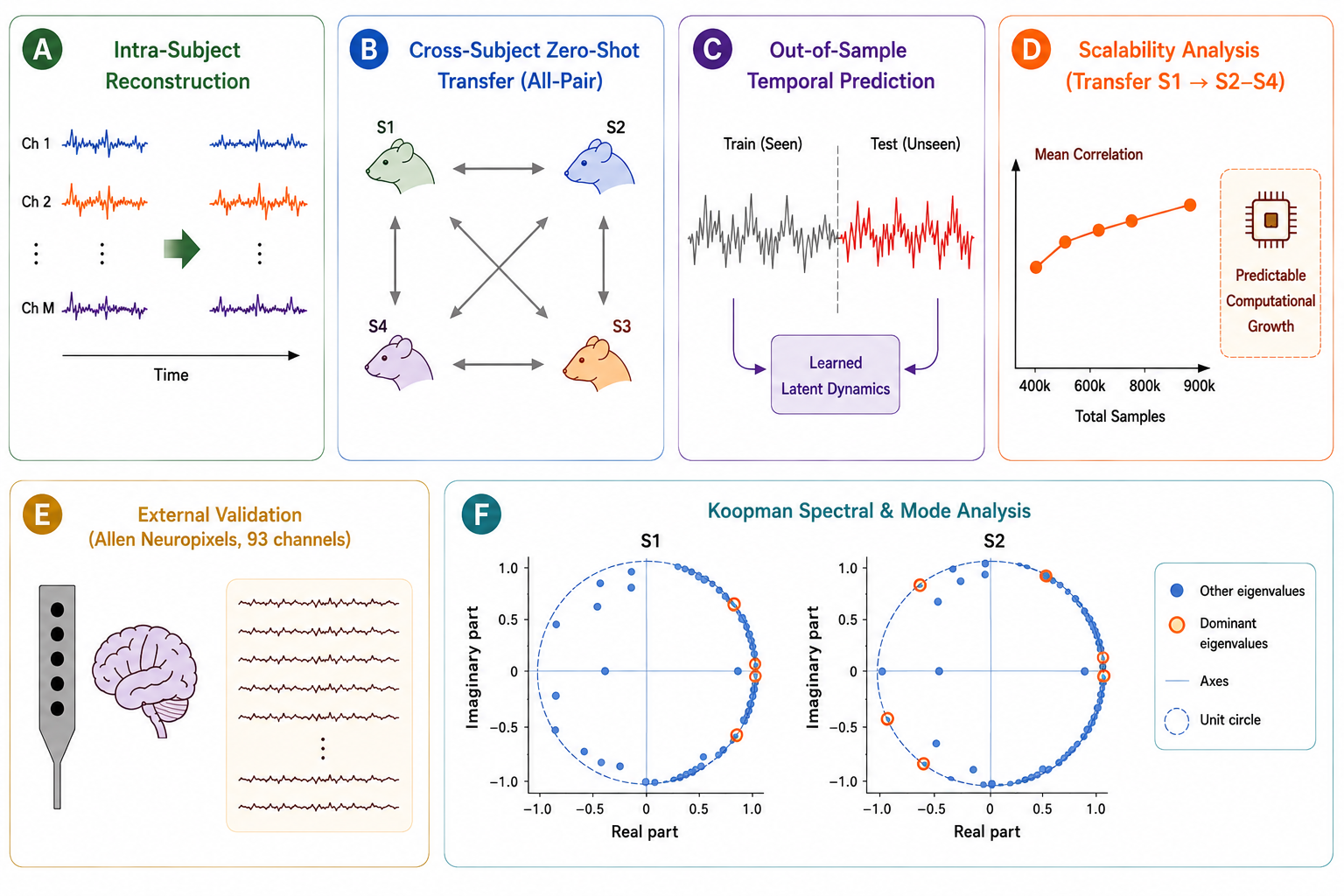}
    \caption{
    Experimental evaluation of PCA-DMD.
    (A) Within-subject reconstruction of hippocampal LFP recordings.
    (B) All-pair cross-subject zero-shot generalization among subjects
    S1--S4 without target-specific fine-tuning.
    (C) The prediction on LFP segments excluded from
    model fitting.
    (D) Scalability analysis from $400{,}000$ to $900{,}000$ samples using
    S1 as the source and S2--S4 as unseen targets.
    (E) External validation on an independent 93-channel Allen Neuropixels
    recording.
    (F) Koopman spectral analysis of the learned latent dynamics through
    eigenvalue distributions in the complex plane.
    }
    \label{fig:evaluation_framework}
\end{figure*}

Figure~\ref{fig:framework} presents the complete PCA-DMD computational
framework. The workflow consists of neural data preparation, overlapping-window
construction, PCA-based latent representation, linear Koopman/DMD evolution,
and reconstruction in the original signal space. The complementary evaluation
settings used to assess reconstruction, cross-subject generalization, prediction, scalability, external validity, and latent
spectral structure are summarized in Figure~\ref{fig:evaluation_framework}.

The framework is applied to two datasets. The primary dataset contains
multichannel hippocampal LFP recordings from four subjects (S1--S4), with
eight channels sampled at $30$ kHz and sequence lengths ranging from
$200{,}000$ to $900{,}000$ samples. An independent Allen Neuropixels LFP
recording with 93 channels and approximately $300{,}000$ samples is used
for external validation.

Each multichannel recording is segmented into overlapping temporal windows
of length $w=3000$ samples with step size $\delta=30$. The windows are
vectorized into high-dimensional snapshot representations and projected
into a compact latent space using PCA. A linear operator is then estimated
in the PCA space to approximate Koopman evolution between consecutive latent
states. The predicted latent states are mapped back to the original signal
space through inverse PCA projection, and the reconstructed windows are
combined using overlap-add aggregation to obtain a continuous multichannel
signal.

The PCA-DMD framework is evaluated under five complementary experimental
settings, as illustrated in Figure~\ref{fig:evaluation_framework}.
First, within-subject reconstruction is evaluated on $200{,}000$-sample
hippocampal recordings and compared with Classical DMD, SpDMD, MrDMD,
and HODMD. Second, cross-subject zero-shot generalization is evaluated at
$300{,}000$ samples using all ordered source--target subject pairs without
target-subject fine-tuning. Third, the prediction is
assessed on temporally held-out LFP segments excluded from model fitting,
providing a direct evaluation of one-step prediction on unseen temporal data.
Fourth, scalability is evaluated from $400{,}000$ to $900{,}000$ samples
by fitting the model on S1 and applying it without fine-tuning to the unseen
target subjects S2--S4. Fifth, the same PCA-DMD reconstruction procedure is
applied to the independent Allen Neuropixels recording to assess robustness
under a different channel dimensionality and acquisition setting.

In addition to reconstruction performance, the learned Koopman operators
are analyzed spectrally by examining their eigenvalue distributions in the
complex plane relative to the unit circle. This analysis characterizes the
latent dynamical structure captured by PCA-DMD.

\subsection{PCA-DMD Reconstruction Framework}

We propose a PCA-DMD method for reconstructing multichannel LFP signals. The raw LFP recordings are first segmented into overlapping windows, then reduced in dimensionality using PCA to obtain compact latent representations. 
A linear Koopman operator (Appendix~\ref{app:koop} ) is subsequently learned in this latent space to capture the temporal evolution of the dynamics across channels. 
Finally, the latent states are mapped back to the original signal space through the inverse PCA projection, and the reconstructed windows are stitched together to approximate the full multichannel LFP signal while preserving SWR dynamics. 

Let the multichannel LFP signal be denoted by $s = \{ s_t \}_{t=1}^T,\, \,  s_t \in \mathbb{R}^m$,
where $m$ is the number of channels and $T$ the total number of samples. We segment the signal into overlapping windows of length \(w\) with step size \(\delta\):
\begin{align*}
X_i &= 
\begin{bmatrix}
s_i^\top & s_{i+1}^\top & \cdots & s_{i+w-1}^\top
\end{bmatrix}^\top 
\in \mathbb{R}^{w \times m}, 
\qquad i = 1, 1+\delta, \cdots .
\end{align*}
Each window is vectorized into \(x_i = \operatorname{vec}(X_i) \in \mathbb{R}^{wm}\), and the \(N\) windows together form the dataset
\begin{align*}
X = 
\begin{bmatrix} 
x_1^\top & x_2^\top & \cdots & x_N^\top 
\end{bmatrix}^\top 
\in \mathbb{R}^{N \times (wm)}.
\end{align*}
We compute a rank-$d$ PCA projection \(W \in \mathbb{R}^{(wm) \times d}\) with \(d \ll wm\).  
The latent representations are
\begin{align*}
Z = X W = 
\begin{bmatrix} 
z_1^\top & z_2^\top & \cdots & z_N^\top 
\end{bmatrix}^\top 
\in \mathbb{R}^{N \times d}, 
\qquad z_i \in \mathbb{R}^d.
\end{align*}
Define the shifted latent pairs as $Z_{\mathrm{past}} = \begin{bmatrix} z_1 & z_2 & \cdots & z_{N-1} \end{bmatrix}$ and $Z_{\mathrm{next}} = \begin{bmatrix} z_2 & z_3 & \cdots & z_N \end{bmatrix}$. We model the temporal evolution in the latent space as $\, z_{i+1} \approx \mathbf{K} z_i, \, \, \mathbf{K} \in \mathbb{R}^{d \times d}$ where the optimal \(\mathbf{K}\) is obtained via least-squares: $\mathbf{K} = \arg\min_{A \in \mathbb{R}^{d \times d}} \| Z_{\mathrm{next}} - A Z_{\mathrm{past}} \|_F 
= Z_{\mathrm{next}} Z_{\mathrm{past}}^\dagger$ with \(\dagger\) the Moore–Penrose pseudoinverse. Then, the predicted latent states evolve as $\, \hat{z}_{i+k} = \mathbf{K}^k z_i$. These are mapped back to the window space via the inverse PCA projection $\hat{x}_i = \hat{z}_i W^\top \in \mathbb{R}^{wm}$ and 
$\hat{X}_i = \operatorname{unvec}(\hat{x}_i) \in \mathbb{R}^{w \times m}$. Finally, the reconstructed multichannel signal \(\hat{s}_t \in \mathbb{R}^m\) is obtained by overlap–add of the windows \(\hat{X}_i\), with a tapering function (e.g., Hann) to reduce boundary artifacts. See Table~\ref{tab:com} for a comparison of our method with other state-of-the-art DMD variants.

\begin{table}[t]
\centering
\caption{DMD variants grouped by SVD workflow and relation to our PCA-DMD.}
\label{tab:DMD_comparison}
\begin{tabular}{p{3cm} p{3cm} p{7cm}}
\toprule
\textbf{Method} & \textbf{Type} & \textbf{Key Notes / Relation} \\
\midrule
Classical DMD \cite{kutz2016DMDclassical} & Snapshot-SVD & 
Koopman regression on snapshot basis. \\
SpDMD \cite{jovanovic2012SparseDMD} & Snapshot-SVD & Adds sparse mode selection; otherwise same as Classical. \\
MrDMD \cite{Kutz2016_2MrDMD} & Window-SVD & Multi-resolution partitioning; trades global coherence for local detail. \\
HODMD \cite{LeClainche2017HODMD} & Window-SVD & Delay embedding; captures high-frequency oscillations but unstable for long windows. \\
Hankel-DMD \cite{hankel} & Window-SVD & Hankel matrix from delays; foundation of HAVOK. \\
HAVOK \cite{brunton2017chaos} & Window-SVD & Hankel embedding + forcing term; chaotic signals. \\
\midrule
\textbf{Our PCA-DMD} & Windowing + PCA & PCA pre-compression + Koopman in latent space + overlap--add reconstruction. Improves stability, scalability, and transient SWR preservation. 
\\
\bottomrule
\end{tabular}\label{tab:com}
\end{table}
\begin{algorithm}[t]
\caption{PCA-DMD for Long-Horizon Reconstruction}
\label{alg:pca_dmd}
\begin{algorithmic}

\State \textbf{Input:} signal $s(t)$, window size $w$, step $\delta$, latent dimension $d$
\State Extract overlapping windows $W_i = s[i\delta : i\delta + w]$
\State Vectorize windows: $x_i = \mathrm{vec}(W_i)$
\State Form consecutive window pairs $(X,Y)$
\State Fit PCA and compute latent coordinates $z_i = P^\top x_i$
\State Estimate Koopman operator: $K^\top = Z^\dagger Z_{+}$

\For{each window $x_i$}
    \State Predict latent state: $\hat z_{i+k} = \mathbf{K}^{k} z_i$
    \State Reconstruct window: $\hat x_{i+k} = P\hat z_{i+k} + \mu$
\EndFor

\State Apply overlap-add averaging to obtain $\hat s(t)$
\State \textbf{Output:} reconstructed signal $\hat s(t)$

\end{algorithmic}
\end{algorithm}

\subsection{Cross-Subject Zero-Shot Generalization}

To investigate the cross-subject generalization of the learned dynamical representations, we introduce a cross-subject zero-shot evaluation protocol. For each ordered subject pair $(a,b)$ with $a\neq b$, the PCA-DMD model is trained exclusively on recordings from source subject $a$, and the resulting parameters are directly applied to target subject $b$ without any additional fine-tuning, such that $\hat{s}^{(b)}=f_{\theta_a}\left(s^{(b)}\right)$. This setting evaluates whether the learned latent dynamics capture subject-invariant neural structure rather than memorizing subject-specific characteristics of individual recordings. By considering all ordered source-target pairs across four subjects, we systematically assess reconstruction fidelity and generalization capability under cross-subject distribution shift. Algorithm~\ref{alg:zero_shot} summarizes the complete zero-shot evaluation procedure.

\begin{algorithm}[t]
\caption{Cross-Subject Zero-Shot Generalization}
\label{alg:zero_shot}
\begin{algorithmic}

\State \textbf{Input:} subjects $\{S_1,\dots,S_n\}$, reconstruction model $f_\theta$

\For{each source subject $S_a$}
    \State Train model parameters $\theta_a$ using only $S_a$

    \For{each target subject $S_b$, $b \neq a$}
        \State Reconstruct target signal:
        $\hat s^{(b)} = f_{\theta_a}(s^{(b)})$
        \State Compute reconstruction metrics between
        $s^{(b)}$ and $\hat s^{(b)}$
    \EndFor
\EndFor

\State \textbf{Output:} cross-subject generalization results

\end{algorithmic}
\end{algorithm}

\subsection{Koopman Spectral Analysis}

To further characterize the learned dynamics, we analyze the spectral properties of the latent Koopman operator. Let $K v_k=\lambda_k v_k$ denote the eigenpairs of $K$, where $\lambda_k$ and $v_k$ are the corresponding eigenvalues and eigenvectors. The latent dynamics can then be expressed as $z_t \approx \sum_{k=1}^{r} c_k \lambda_k^t v_k$, such that the eigenvalue spectrum provides insight into the stability, persistence, and oscillatory behavior of the reconstructed dynamics, while the corresponding Koopman modes reveal the spatiotemporal structures captured by the latent representation. The PCA-induced low-rank structure further improves computational scalability: given $X=U\Sigma V^\top$, the rank-$r$ approximation $X_r=U_r\Sigma_rV_r^\top$ retains the dominant variance while substantially reducing the dimensionality of the original signal space. This compression enables efficient Koopman regression, improves numerical conditioning, and facilitates stable long-horizon reconstruction of large-scale multichannel neural recordings.

\section{Results}
\label{sec:results}

We evaluated PCA-DMD under five complementary experimental settings.
First, we assessed within-subject reconstruction of $200{,}000$-sample
hippocampal LFP recordings and compared the proposed method with
conventional DMD variants. Second, we investigated cross-subject
zero-shot generalization using $300{,}000$ samples across all ordered
source--target pairs among four subjects, without target-subject
fine-tuning. Third, we evaluated prediction on
temporally held-out LFP segments excluded from model fitting. Fourth,
we performed a dedicated scalability analysis in which a PCA-DMD model
fitted on source subject S1 was applied without fine-tuning to the unseen
target subjects S2, S3, and S4 for signal lengths ranging from
$400{,}000$ to $900{,}000$ samples. Finally, we evaluated the framework
on an independent 93-channel Allen Neuropixels LFP recording. Together,
these experiments assess reconstruction fidelity, cross-subject
generalization, prediction, long-horizon
scalability, and robustness to an independent acquisition setting.

\subsection{Within-Subject Reconstruction at 200k Samples}
\label{sec:results_200k}

We first evaluated whether PCA-DMD could reconstruct long multichannel
hippocampal LFP segments while preserving the transient oscillatory
structure associated with sharp wave-ripples. All methods were evaluated
on the same $200{,}000$-sample recording using identical signal
preprocessing and reconstruction settings.

\begin{figure*}[t]
    \centering
    \includegraphics[width=.84\textwidth]{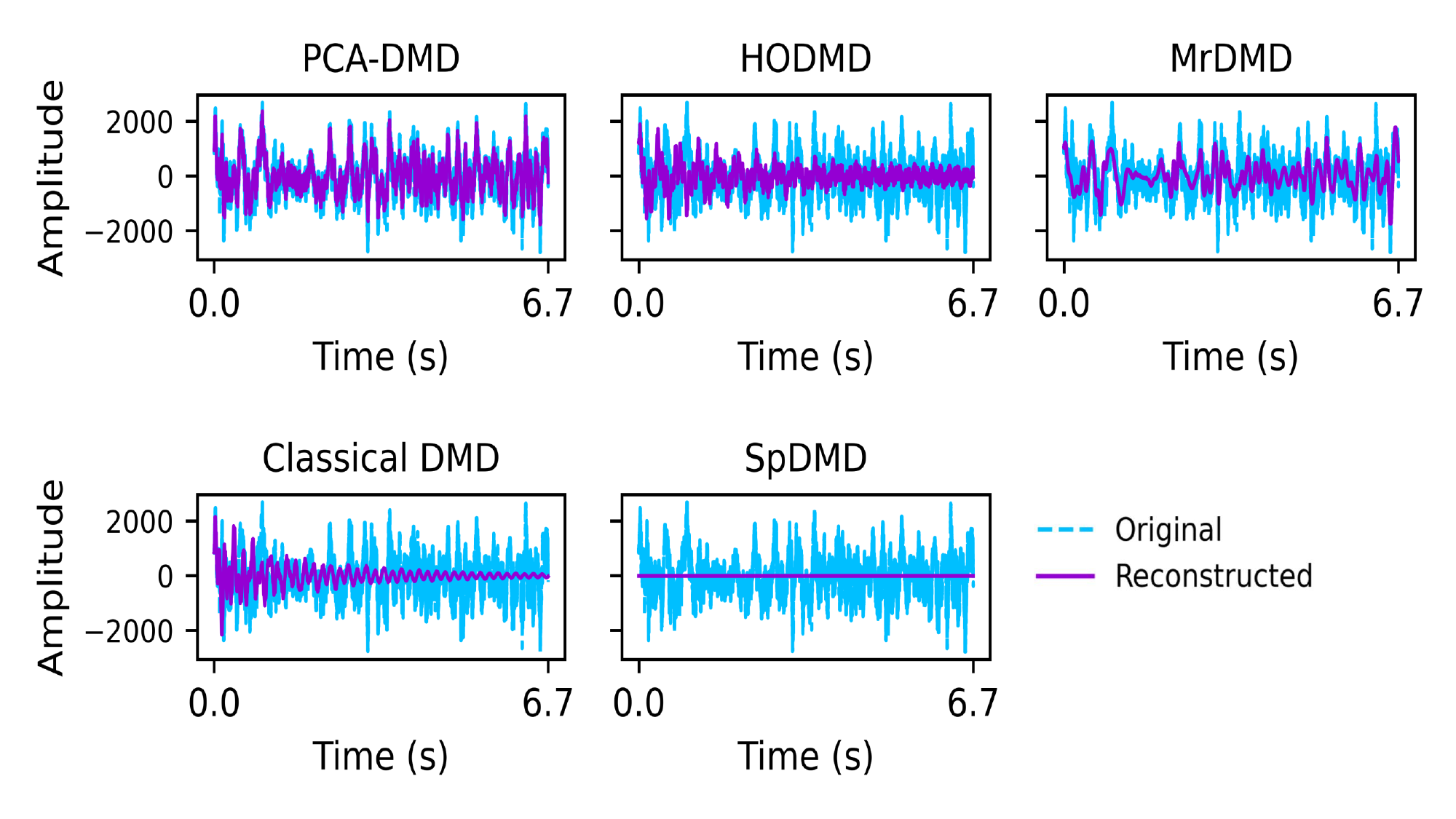}
    \caption{
    Within-subject reconstruction of a $200{,}000$-sample hippocampal
    LFP segment. Original and reconstructed signals are shown for
    PCA-DMD and the conventional DMD variants. PCA-DMD most closely
    follows the original waveform and better preserves transient
    oscillatory structure, whereas the baseline methods show varying
    degrees of attenuation, smoothing, and temporal distortion.
    }
    \label{fig:results_200k_overlay}
\end{figure*}

Figure~\ref{fig:results_200k_overlay} compares representative
reconstructions obtained using PCA-DMD, Classical DMD, HODMD, MrDMD,
and SpDMD. PCA-DMD produced the closest visual agreement with the
original signal. Both the slowly varying waveform structure and the
faster amplitude fluctuations were retained over the reconstruction
interval. In contrast, the conventional DMD variants exhibited different
forms of degradation, including amplitude attenuation, excessive
smoothing, temporal misalignment, and loss of high-frequency structure.

\begin{figure*}[t]
    \centering
    \includegraphics[width=.84\textwidth]{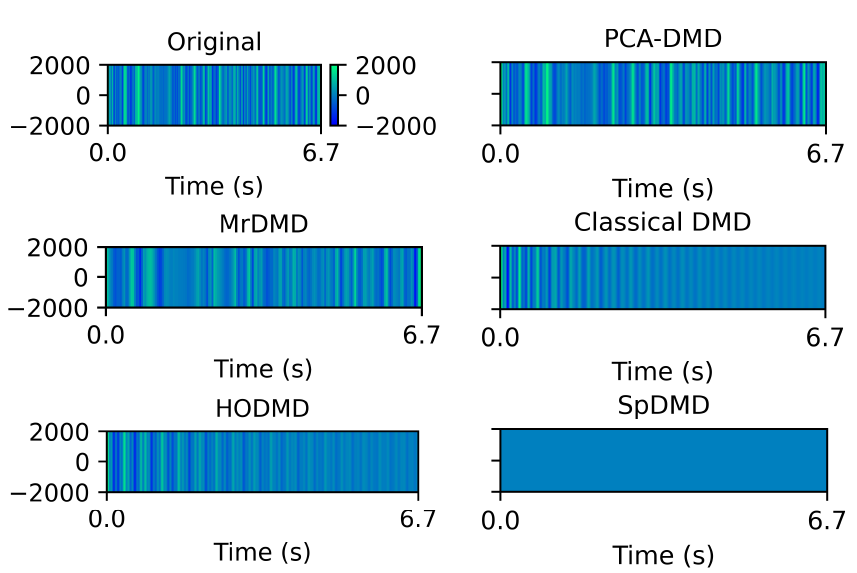}
    \caption{
    Multichannel heatmap comparison for the $200{,}000$-sample
    reconstruction experiment. The PCA-DMD reconstruction preserves the
    channel-wise and temporal amplitude organization of the original
    recording more faithfully than the conventional DMD variants.
    }
    \label{fig:results_200k_heatmaps}
\end{figure*}

The multichannel heatmaps in
Figure~\ref{fig:results_200k_heatmaps} provide a complementary view of
the reconstructed spatiotemporal organization. The PCA-DMD heatmap
closely reproduced the temporal amplitude patterns observed across the
eight original channels. MrDMD retained part of the coarse temporal
organization but lost finer detail, whereas Classical DMD, HODMD, and
SpDMD produced substantially more homogeneous or distorted
representations. These observations indicate that the advantage of
PCA-DMD is not restricted to a selected channel but extends across the
multichannel signal.

The quantitative comparison further supports the visual findings.
PCA-DMD achieved a KLD of $0.0761$ and an HD of $0.0847$, the lowest
values among the evaluated methods. The second-best results were obtained
by MrDMD, with a KLD of $0.4852$ and an HD of $0.2238$. Classical DMD
and HODMD showed larger distributional discrepancies, with KLD values
of $0.8710$ and $0.6135$ and HD values of $0.4067$ and $0.3050$,
respectively. SpDMD exhibited the largest discrepancies, with a KLD of
$14.4061$ and an HD of $0.9074$. Thus, the quantitative results are
consistent with the waveform and multichannel comparisons and show that
performing Koopman regression in a PCA-compressed latent space
substantially improves reconstruction fidelity relative to direct
application of the conventional DMD formulations.

\subsection{Cross-Subject Zero-Shot Generalization at 300k Samples}
\label{sec:results_300k}

We next evaluated whether the latent dynamical representation learned by
PCA-DMD generalizes across subjects. For each ordered source--target pair,
the PCA basis and latent Koopman operator were fitted exclusively on the
source subject and then applied directly to the unseen target subject
without target-specific fine-tuning. Across subjects S1--S4, this
resulted in 12 ordered zero-shot generalization experiments.

\begin{figure*}[t]
    \centering
    \includegraphics[width=.96\textwidth]{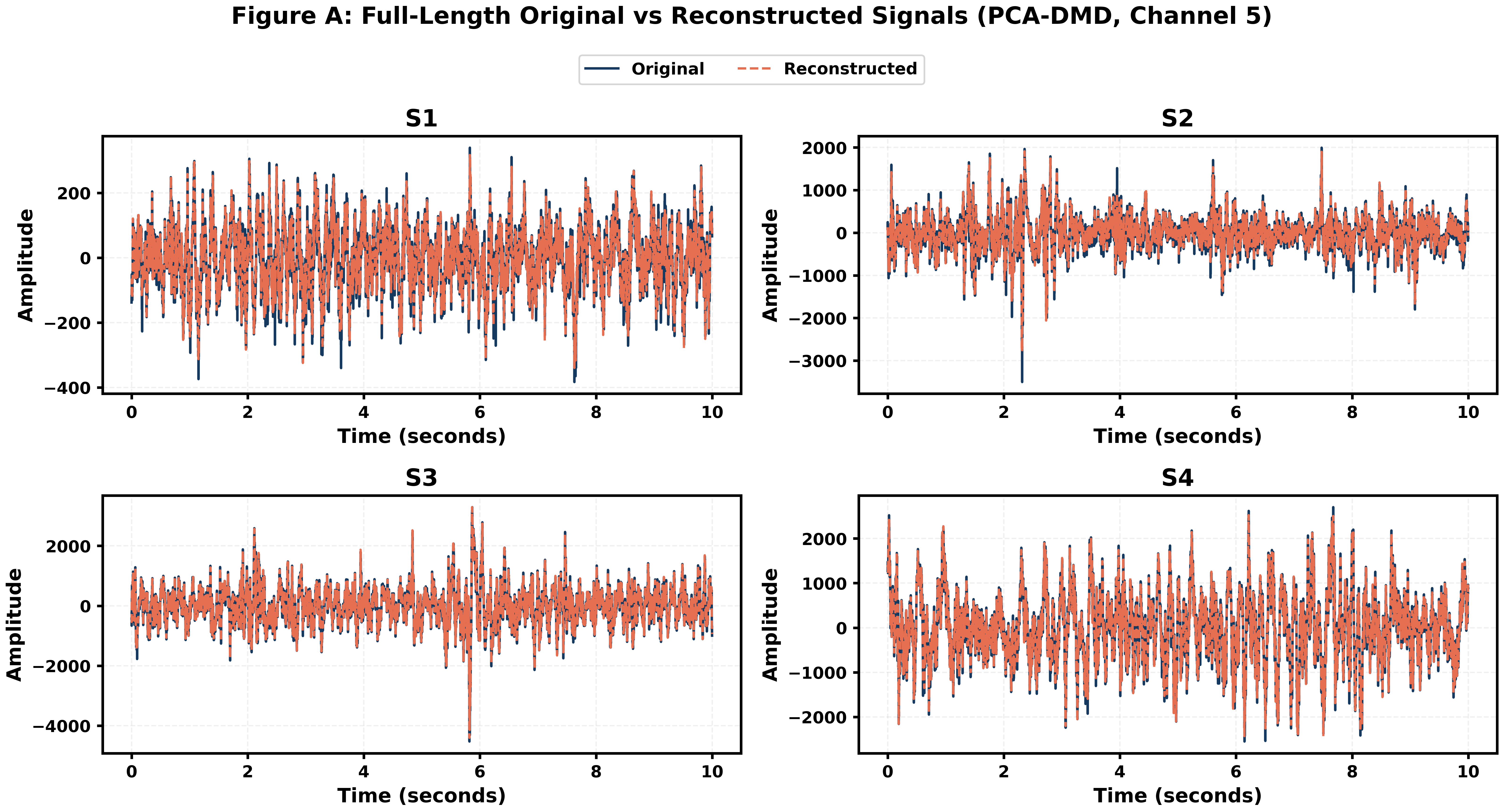}
    \caption{
    Representative PCA-DMD reconstructions from the $300{,}000$-sample
    cross-subject zero-shot generalization experiment. Full-length and
    zoomed overlays are shown for Channel 5 across the four subjects.
    Solid curves denote the original target signal and dashed curves
    denote the PCA-DMD reconstruction.
    }
    \label{fig:results_300k_reconstruction}
\end{figure*}

Representative full-length and locally zoomed reconstructions are shown
in Figure~\ref{fig:results_300k_reconstruction}. At the full-signal
scale, the reconstructed trajectories closely followed the temporal
envelope and amplitude variation of the unseen target recordings. The
zoomed panels further show that PCA-DMD recovered local waveform
evolution rather than reproducing only a smoothed low-frequency trend.
This agreement was observed across target subjects despite differences
in signal amplitude and local waveform morphology.

Across all ordered source--target pairs, PCA-DMD achieved correlations
between $0.9504$ and $0.9800$, HD values between $0.0010$ and $0.0072$,
and KLD values between $0.0005$ and $0.0022$. The consistently high
correlations indicate strong temporal agreement on unseen target subjects,
while the small HD and KLD values indicate close agreement between the
original and reconstructed target-subject amplitude distributions. These
results demonstrate strong cross-subject zero-shot generalization without
target-specific fine-tuning.

The conventional DMD variants degraded substantially under the same
cross-subject evaluation. Classical DMD, HODMD, and HankelDMD produced
correlations close to $0.10$, with HD values of approximately
$0.10$--$0.26$ and KLD values of approximately $3$--$5$. SpDMD showed
a modest improvement over these methods but remained considerably below
PCA-DMD, with correlations of approximately $0.15$--$0.27$ and KLD
values of approximately $1.7$--$3.2$. The large performance gap indicates
that the cross-subject generalization observed with PCA-DMD is not
reproduced by the conventional DMD variants under the same zero-shot
setting.

\begin{table}[t]
    \centering
    \caption{
    Summary of the $300{,}000$-sample cross-subject zero-shot
    generalization results. PCA-DMD values represent the range across
    all ordered source--target subject pairs. Baseline values summarize
    the observed operating ranges. Higher correlation is better; lower
    HD and KLD are better.
    }
    \label{tab:results_300k_zero_shot}
    \begin{tabular}{lccc}
        \toprule
        Method
        & Corr $\uparrow$
        & HD $\downarrow$
        & KLD $\downarrow$ \\
        \midrule
        PCA-DMD
        & 0.9504--0.9800
        & 0.0010--0.0072
        & 0.0005--0.0022 \\
        Classical DMD
        & $\sim$0.10
        & $\sim$0.10--0.26
        & $\sim$3--5 \\
        HODMD
        & $\sim$0.10
        & $\sim$0.10--0.26
        & $\sim$3--5 \\
        HankelDMD
        & $\sim$0.10
        & $\sim$0.10--0.26
        & $\sim$3--5 \\
        SpDMD
        & $\sim$0.15--0.27
        & $\sim$0.07--0.25
        & $\sim$1.7--3.2 \\
        \bottomrule
    \end{tabular}
\end{table}

\begin{figure*}[t]
    \centering
    \includegraphics[width=.93\textwidth]{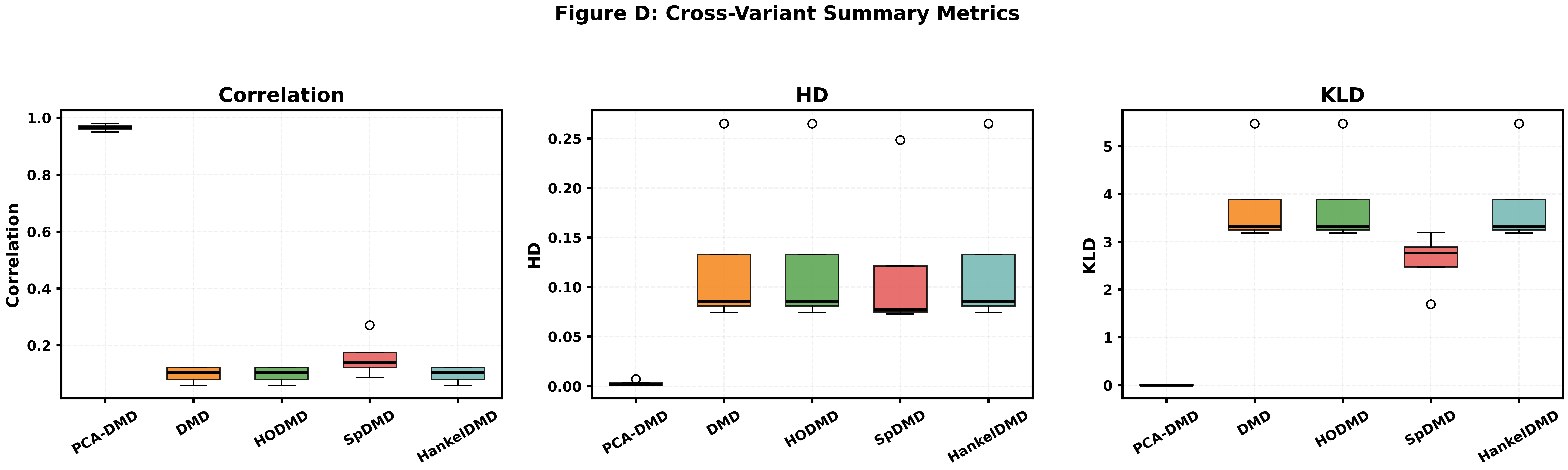}
    \caption{
    Cross-subject zero-shot generalization performance at $300{,}000$
    samples. The boxplots compare correlation, HD, and KLD across
    methods, while the source-to-target heatmaps show the PCA-DMD
    results for all ordered subject pairs.
    }
    \label{fig:results_300k_metrics}
\end{figure*}

Figure~\ref{fig:results_300k_metrics} summarizes the metric
distributions across methods. PCA-DMD forms a clearly separated
high-correlation regime while simultaneously yielding the smallest HD
and KLD values across the ordered source--target pairs. This separation
provides further evidence that the learned PCA-DMD representation
generalizes across subjects under the zero-shot evaluation setting.


\subsection{Zero-Shot Scalability from 400k to 900k Samples}
\label{sec:results_scalability}

The preceding experiments established reconstruction performance at
$200{,}000$ samples and cross-subject zero-shot generalization at
$300{,}000$ samples. We next examined whether PCA-DMD maintains stable
performance as the signal length increases substantially beyond these
operating ranges. The scalability analysis considered signal lengths of

\[
400k,\ 500k,\ 600k,\ 700k,\ 800k,\ \text{and}\ 900k.
\]

At each signal length, a new PCA-DMD model was fitted exclusively on
source subject S1 and then applied without fine-tuning to the unseen
target subjects S2, S3, and S4. The model was fitted independently at
each signal length rather than initialized from a model obtained at a
shorter horizon. Window length, step size, latent dimension, histogram
construction, and metric definitions were held fixed across all
experiments. This design isolates the effect of increasing signal length
while retaining the same zero-shot source-to-target evaluation setting.

\begin{figure*}[t]
    \centering
    \includegraphics[width=.90\textwidth]{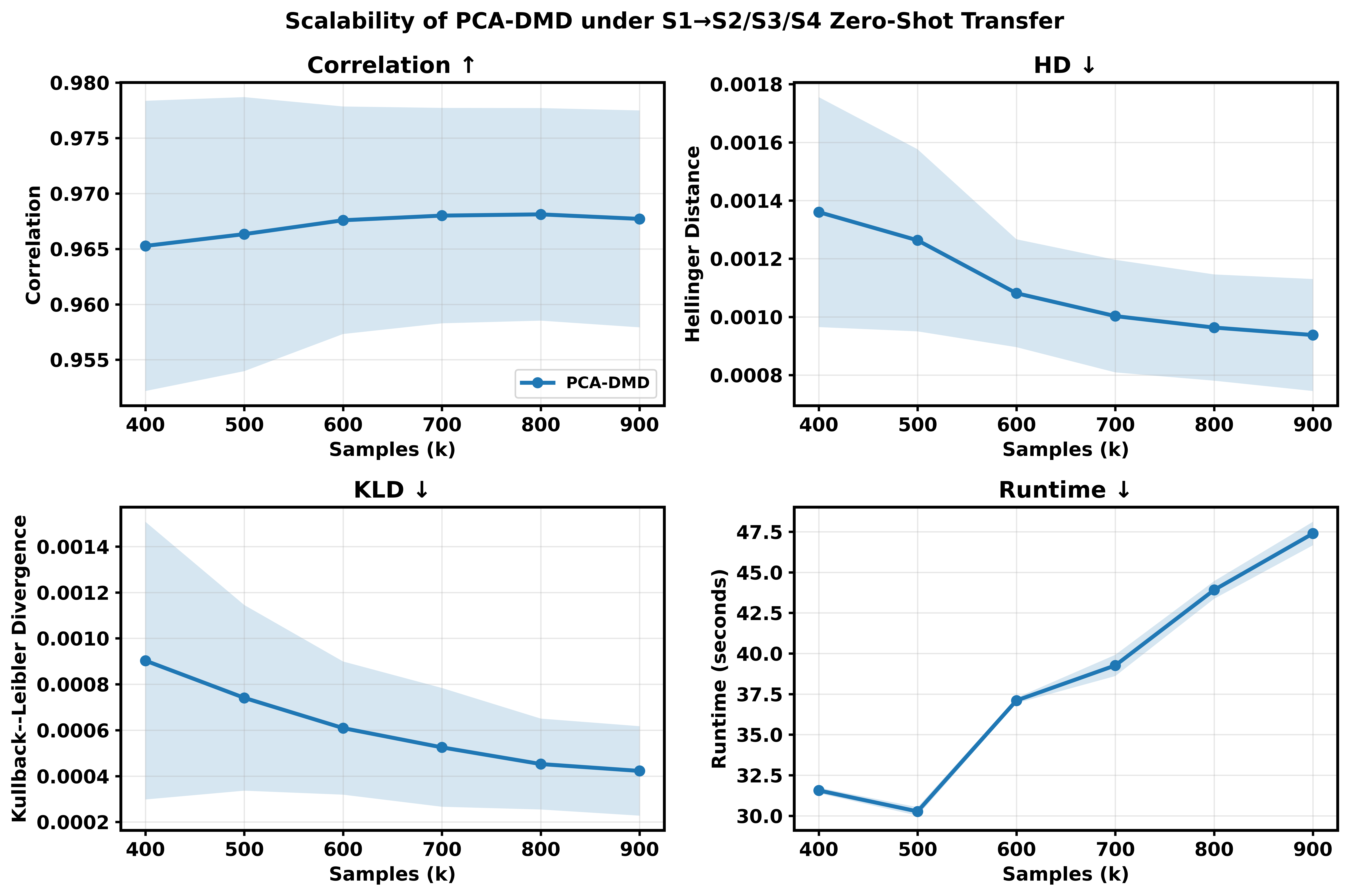}
    \caption{
    Scalability of PCA-DMD under S1$\rightarrow$S2/S3/S4 zero-shot
    evaluation. Curves show the mean performance across the three unseen
    target subjects, and shaded regions indicate variability across
    S2--S4. Correlation remains stable from $400k$ to $900k$ samples,
    while HD and KLD decrease. Runtime increases with signal length.
    }
    \label{fig:results_scalability_metrics}
\end{figure*}

Figure~\ref{fig:results_scalability_metrics} shows the mean performance
across the three unseen target subjects. Mean correlation remained highly
stable, increasing from $0.96527$ at $400k$ samples to $0.96812$ at
$800k$ samples and remaining at $0.96771$ at $900k$. Thus, increasing
the sequence length by more than a factor of two did not produce
cumulative degradation in reconstruction agreement.

Distributional agreement also remained stable and improved with increasing
signal length. Mean HD decreased from $1.360\times10^{-3}$ at $400k$ to
$9.377\times10^{-4}$ at $900k$, while mean KLD decreased from
$9.033\times10^{-4}$ to $4.223\times10^{-4}$. Mean NRMSE remained within
a narrow range, decreasing from $0.2585$ at $400k$ to approximately
$0.249$--$0.251$ at the longer signal lengths.

\begin{figure*}[t]
    \centering
    \includegraphics[width=.95\textwidth]{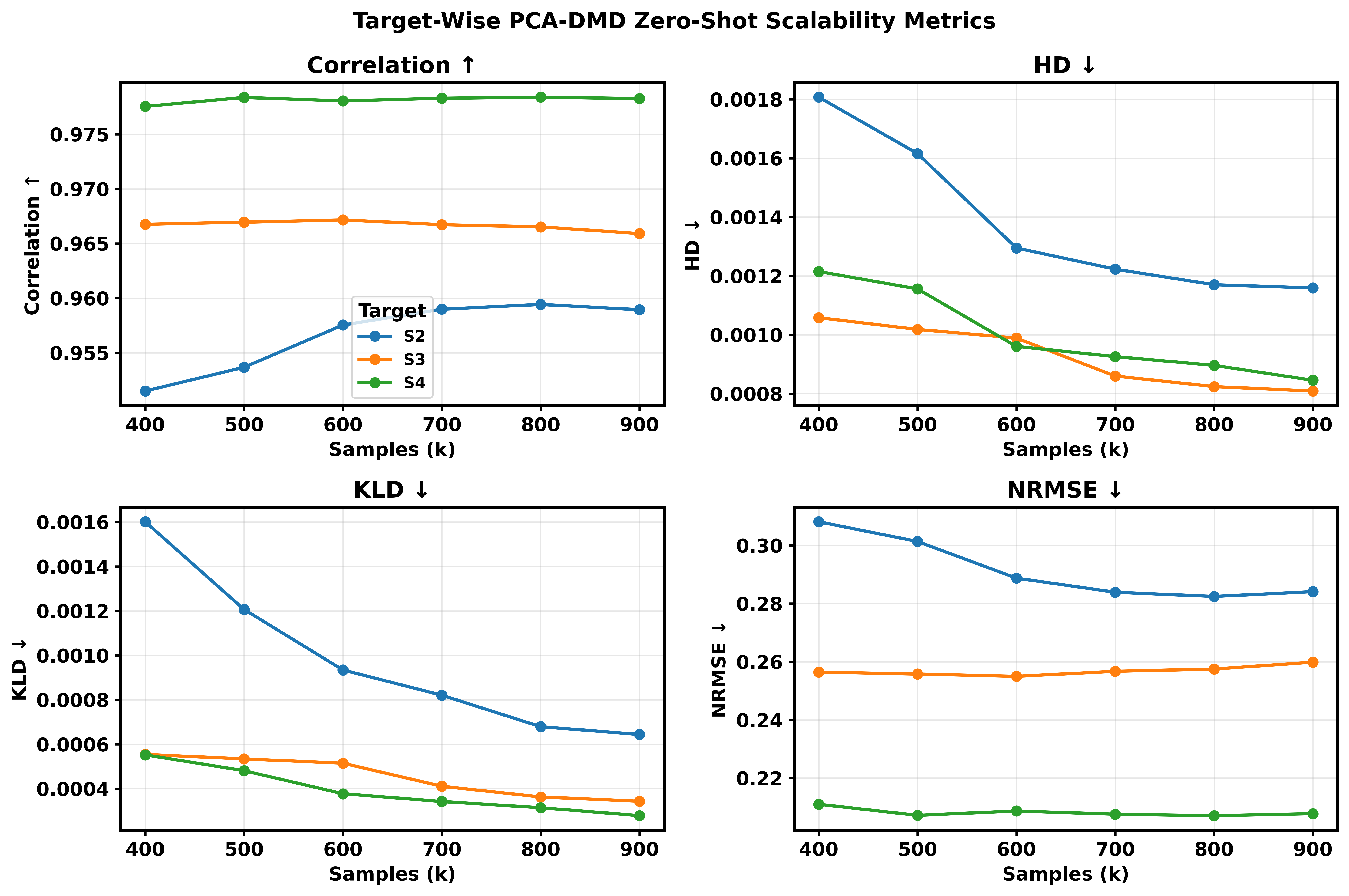}
    \caption{
    Target-wise PCA-DMD scalability results. Correlation, HD, KLD, and
    NRMSE are shown separately for S1$\rightarrow$S2,
    S1$\rightarrow$S3, and S1$\rightarrow$S4. Stable performance across
    increasing signal lengths is observed consistently for all three
    unseen target subjects.
    }
    \label{fig:results_scalability_targets}
\end{figure*}

The target-wise results in
Figure~\ref{fig:results_scalability_targets} show that the aggregate
behavior was not driven by a single target subject. For S2, correlation
increased from $0.95150$ at $400k$ to a maximum of $0.95942$ at $800k$
and remained at $0.95895$ at $900k$. Over the same range, HD decreased
from $0.001807$ to $0.001159$, and KLD decreased from $0.001602$ to
$0.000644$.

Performance on S3 remained nearly constant, with correlations between
$0.96591$ and $0.96716$. Its HD decreased from $0.001058$ to
$0.000809$, while KLD decreased from $0.000555$ to $0.000344$.
S4 consistently achieved the strongest reconstruction performance, with
correlations between $0.97755$ and $0.97840$. At $900k$ samples, S4
achieved an HD of $0.000845$, a KLD of $0.000279$, and an NRMSE of
$0.20772$.

\begin{figure*}[t]
    \centering
    \includegraphics[width=.90\textwidth]{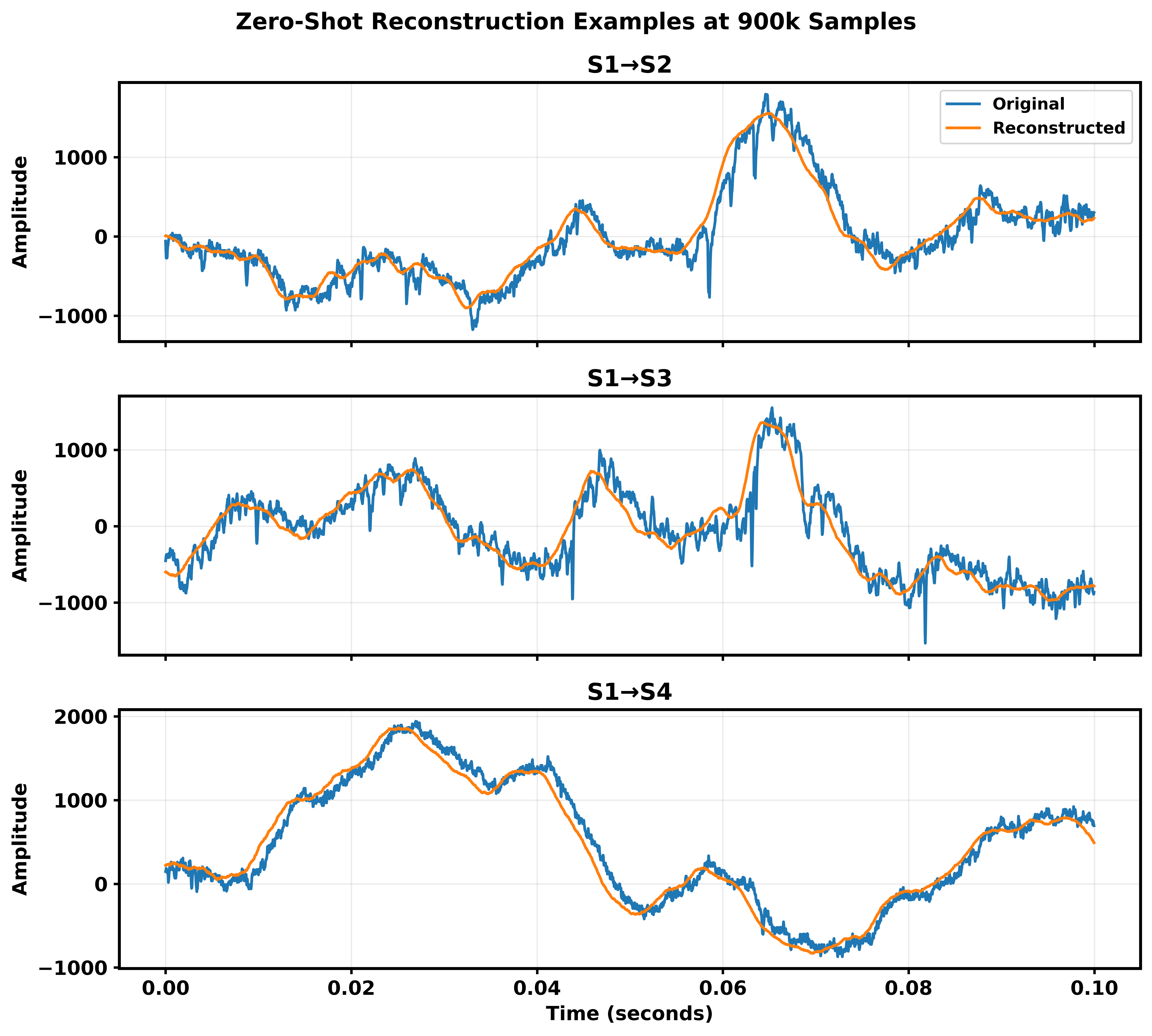}
    \caption{
    Full-length PCA-DMD zero-shot reconstructions at $900{,}000$
    samples for S1$\rightarrow$S2, S1$\rightarrow$S3, and
    S1$\rightarrow$S4. Channel 5 is shown for visual clarity. The
    selected zoomed panel illustrates local waveform agreement.
    }
    \label{fig:results_900k_overlays}
\end{figure*}

The full-length $900k$-sample reconstructions are shown in
Figure~\ref{fig:results_900k_overlays}. At the original sampling rate of
$30$ kHz, this corresponds to approximately $30$ seconds of continuous
multichannel LFP activity. PCA-DMD preserved the signal envelope and
major transient fluctuations throughout the full interval for all three
unseen target subjects. The zoomed reconstruction further demonstrates
close local alignment between the original and reconstructed waveforms.

\begin{figure*}[t]
    \centering
    \includegraphics[width=.95\textwidth]{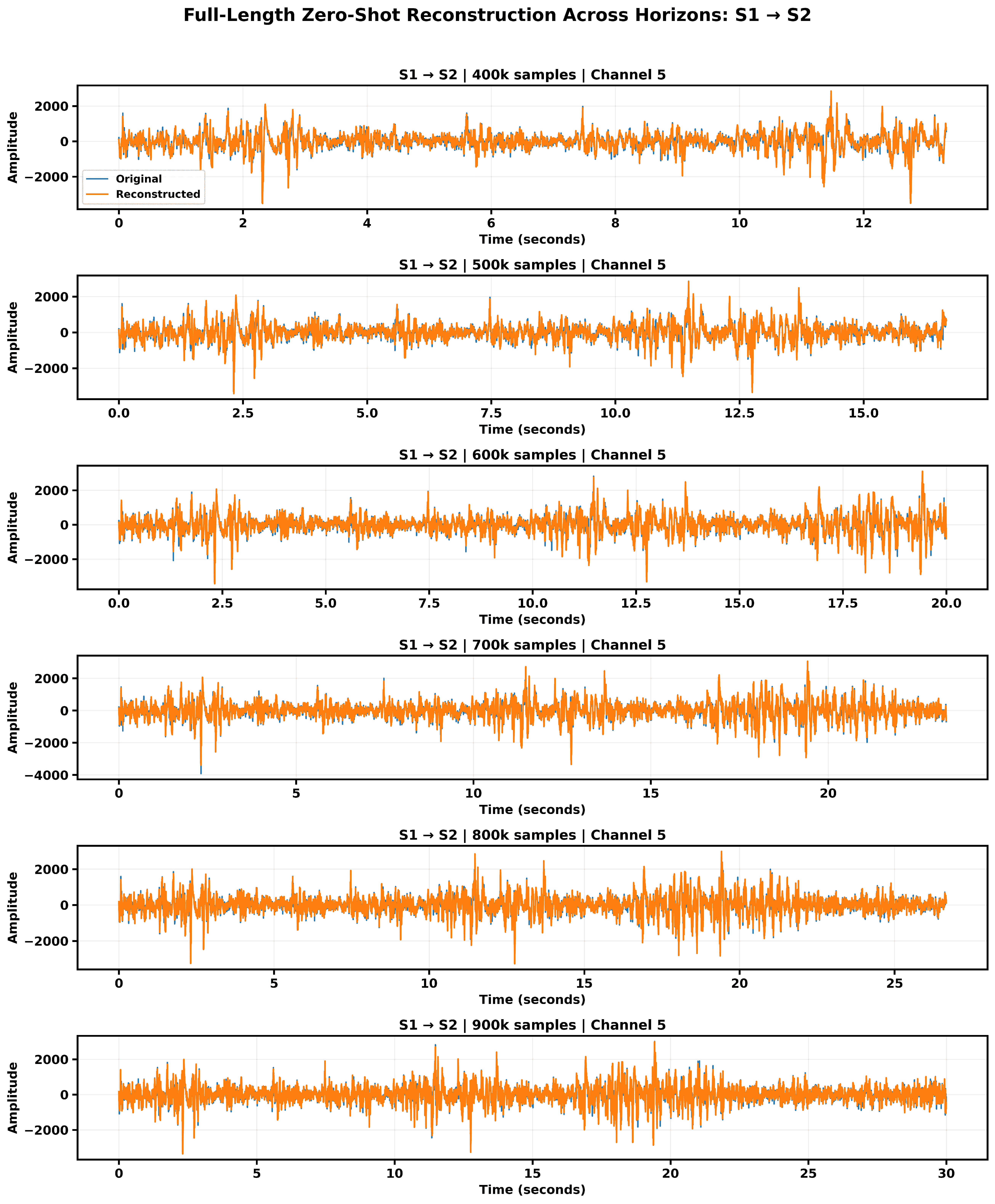}
    \caption{
    Full-length PCA-DMD zero-shot reconstructions for
    S1$\rightarrow$S2 across signal lengths from $400{,}000$ to
    $900{,}000$ samples. Channel 5 is shown to illustrate reconstruction
    behavior as the signal length increases.
    }
    \label{fig:results_all_horizons_overlay}
\end{figure*}

Figure~\ref{fig:results_all_horizons_overlay} further illustrates the
full-length reconstruction behavior across the complete range from
$400{,}000$ to $900{,}000$ samples, providing a direct visual comparison
of reconstruction stability as the signal length increases.

Computational cost increased predictably with signal length. The average
total runtime across target subjects was approximately $31.56$ seconds
at $400k$ samples and $47.39$ seconds at $900k$ samples. At $900k$,
model fitting required $42.23$ seconds, while target reconstruction
required between $4.73$ and $6.00$ seconds. Thus, most of the
computational cost was associated with fitting the source representation,
whereas direct application to an unseen target subject remained
comparatively inexpensive. Overall, these results demonstrate that
PCA-DMD maintains stable zero-shot reconstruction performance as the
signal length increases from $400k$ to $900k$ samples.

\begin{table*}[t]
    \centering
    \caption{
    Summary of the S1-based zero-shot scalability experiment. Metric
    ranges are taken across the six signal lengths from $400k$ to
    $900k$. Higher correlation is better; lower HD, KLD, and NRMSE are
    better.
    }
    \label{tab:results_scalability_summary}
    \begin{tabular}{lcccc}
        \toprule
        Target
        & Corr $\uparrow$
        & HD $\downarrow$
        & KLD $\downarrow$
        & NRMSE $\downarrow$ \\
        \midrule
        S2
        & 0.95150--0.95942
        & 0.001159--0.001807
        & 0.000644--0.001602
        & 0.28247--0.30816 \\
        S3
        & 0.96591--0.96716
        & 0.000809--0.001058
        & 0.000344--0.000555
        & 0.25498--0.25984 \\
        S4
        & 0.97755--0.97840
        & 0.000845--0.001215
        & 0.000279--0.000553
        & 0.20705--0.21099 \\
        \bottomrule
    \end{tabular}
\end{table*}

A compact summary is provided in
Table~\ref{tab:results_scalability_summary}. The complete target-wise
table, including training time, prediction time, MSE, RMSE, MAE, and
NRMSE at every horizon, is reported in Table ~\ref{tab:scalability_zero_shot}.

\begin{table*}[t]
\centering
\caption{Scalability analysis of zero-shot cross-subject transfer from source
subject S1 to target subjects S2--S4. Training is performed independently at
each signal length, without loading a pretrained checkpoint. Correlation
(Corr) is higher-is-better, whereas Hellinger distance (HD), Kullback--Leibler
divergence (KLD), MSE, RMSE, MAE, and normalized RMSE (NRMSE) are
lower-is-better. Bold values indicate the best reconstruction result within
each target subject.}
\label{tab:scalability_zero_shot}
\scriptsize
\setlength{\tabcolsep}{3.2pt}
\renewcommand{\arraystretch}{1.08}

\resizebox{\textwidth}{!}{%
\begin{tabular}{ccrrrrrrrrrr}
\toprule
\multirow{2}{*}{Target}
& \multirow{2}{*}{Samples ($\times 10^{3}$)}
& \multicolumn{3}{c}{Runtime (s)}
& \multicolumn{7}{c}{Reconstruction metrics} \\
\cmidrule(lr){3-5}
\cmidrule(lr){6-12}
& & Train & Prediction & Total
& Corr $\uparrow$
& HD $\downarrow$
& KLD $\downarrow$
& MSE $\downarrow$
& RMSE $\downarrow$
& MAE $\downarrow$
& NRMSE $\downarrow$ \\
\midrule

\multirow{6}{*}{S2}
& 400 & 29.32 & 2.40 & 31.73
& 0.95150 & 0.001807 & 0.001602
& 23411.85 & 153.01 & 102.94 & 0.30816 \\

& 500 & 27.56 & 2.97 & 30.53
& 0.95368 & 0.001616 & 0.001207
& 21119.21 & 145.32 & 98.76 & 0.30137 \\

& 600 & 34.01 & 3.29 & 37.29
& 0.95754 & 0.001295 & 0.000935
& 24849.90 & 157.64 & 105.77 & 0.28877 \\

& 700 & 35.22 & 4.80 & 40.02
& 0.95899 & 0.001223 & 0.000821
& 24279.62 & 155.82 & 104.84 & 0.28391 \\

& 800 & 39.51 & 5.01 & 44.52
& \textbf{0.95942} & 0.001170 & 0.000679
& 21900.50 & 147.99 & 99.67 & \textbf{0.28247} \\

& 900 & 42.23 & 6.00 & 48.23
& 0.95895 & \textbf{0.001159} & \textbf{0.000644}
& \textbf{20865.32} & \textbf{144.45} & \textbf{97.83} & 0.28411 \\

\midrule

\multirow{6}{*}{S3}
& 400 & 29.32 & 2.06 & 31.38
& 0.96676 & 0.001058 & 0.000555
& \textbf{23484.25} & \textbf{153.25} & 112.14 & 0.25646 \\

& 500 & 27.56 & 2.70 & 30.26
& 0.96695 & 0.001018 & 0.000534
& 23579.13 & 153.55 & \textbf{111.83} & 0.25577 \\

& 600 & 34.01 & 3.03 & 37.03
& \textbf{0.96716} & 0.000989 & 0.000515
& 23937.28 & 154.72 & 112.58 & \textbf{0.25498} \\

& 700 & 35.22 & 3.66 & 38.88
& 0.96673 & 0.000860 & 0.000411
& 26366.42 & 162.38 & 115.59 & 0.25671 \\

& 800 & 39.51 & 4.25 & 43.76
& 0.96653 & 0.000824 & 0.000363
& 26225.52 & 161.94 & 116.47 & 0.25749 \\

& 900 & 42.23 & 4.73 & 46.96
& 0.96591 & \textbf{0.000809} & \textbf{0.000344}
& 28158.59 & 167.81 & 121.75 & 0.25984 \\

\midrule

\multirow{6}{*}{S4}
& 400 & 29.32 & 2.25 & 31.58
& 0.97755 & 0.001215 & 0.000553
& 23373.36 & 152.88 & 113.00 & 0.21099 \\

& 500 & 27.56 & 2.45 & 30.01
& 0.97837 & 0.001156 & 0.000481
& 23476.90 & 153.22 & 112.87 & 0.20715 \\

& 600 & 34.01 & 2.98 & 36.98
& 0.97805 & 0.000961 & 0.000378
& 25058.98 & 158.30 & 116.01 & 0.20866 \\

& 700 & 35.22 & 3.67 & 38.89
& 0.97830 & 0.000926 & 0.000343
& 24065.31 & 155.13 & 114.02 & 0.20753 \\

& 800 & 39.51 & 3.95 & 43.47
& \textbf{0.97840} & 0.000896 & 0.000315
& 22870.83 & 151.23 & 111.35 & \textbf{0.20705} \\

& 900 & 42.23 & 4.76 & 46.99
& 0.97826 & \textbf{0.000845} & \textbf{0.000279}
& \textbf{22379.89} & \textbf{149.60} & \textbf{110.13} & 0.20772 \\

\bottomrule
\end{tabular}%
}
\end{table*}

\subsection{Prediction}
\label{sec:temporal_prediction}

Beyond reconstruction on the evaluation interval, we examined whether the
latent dynamics learned by PCA-DMD can predict temporally unseen LFP states.
This experiment evaluates \emph{prediction}
rather than autonomous long-horizon forecasting. Specifically, PCA and the
DMD operator are fitted exclusively on the designated training segment,
while the subsequent validation interval is completely excluded from model
fitting. The learned PCA basis and latent evolution operator are then fixed
and applied to this unseen temporal segment. Let $\mathbf{x}_t$ denote a
high-dimensional LFP snapshot and $\mathbf{z}_t$ its representation in the
PCA latent space. For an unseen validation state, one-step latent prediction
is computed using the operator learned exclusively from the training data as
$\hat{\mathbf{z}}_{t+1}=\mathbf{K}\mathbf{z}_t$, followed by inverse PCA
projection as
$\hat{\mathbf{x}}_{t+1}=\mathbf{P}^{\dagger}\hat{\mathbf{z}}_{t+1}$,
where $\mathbf{K}$ is the fixed DMD operator and
$\mathbf{P}^{\dagger}$ denotes the inverse mapping from the PCA latent space
to the original signal space. Thus, each prediction evaluates whether the
learned latent transition law can predict the next state from states
encountered in a temporally distinct and previously unseen portion of the
recording. Importantly, the model is not recursively propagated from its own
previous predictions over the entire validation interval; therefore, this
experiment measures one-step prediction rather than
autonomous long-horizon forecasting.

\begin{figure*}[!t]
    \centering
    \includegraphics[width=\textwidth]{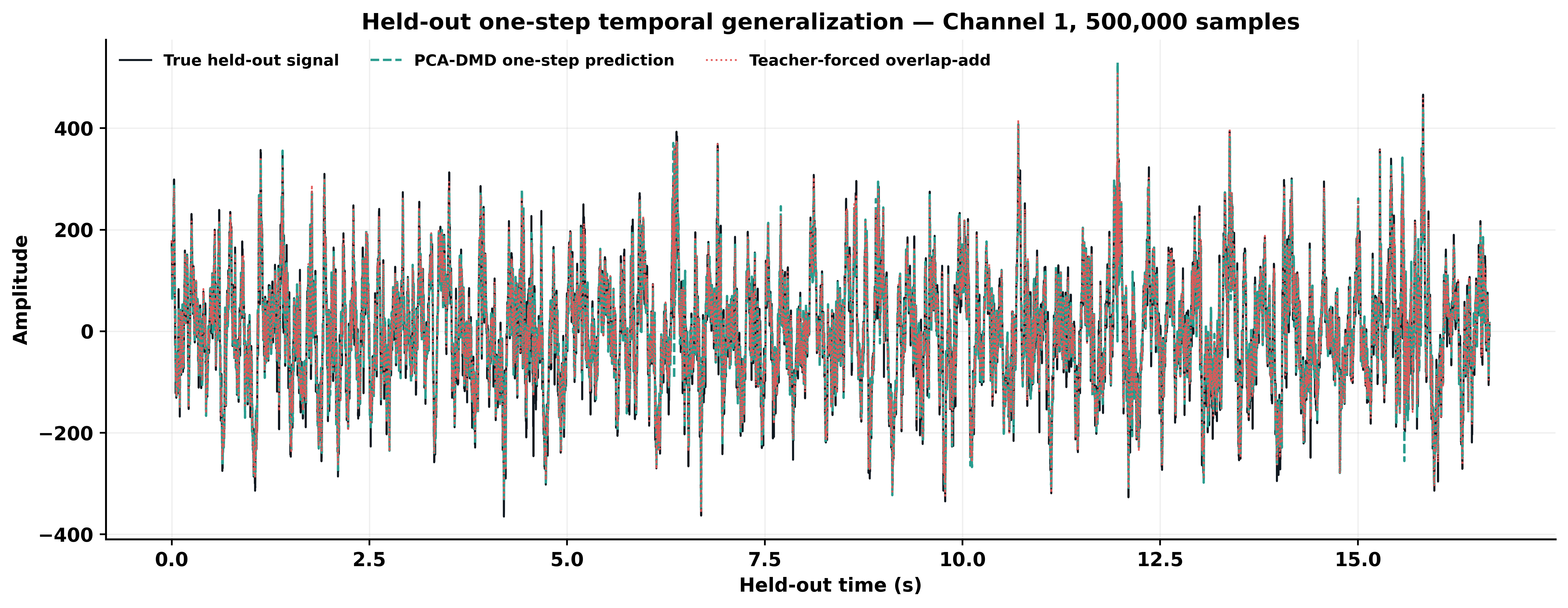}
    \caption{
    The prediction on the held-out LFP interval. The true
    signal and PCA-DMD one-step predictions are shown for a representative
    channel using a model fitted exclusively on the training segment. Close
    agreement across the unseen interval demonstrates accurate prediction of
    temporally held-out neural activity.
    }
    \label{fig:temporal_prediction_full}
\end{figure*}

Figure~\ref{fig:temporal_prediction_full} shows the one-step predictions
across the complete held-out interval for a representative channel. The
PCA-DMD predictions closely follow the temporal structure and amplitude
variations of the unseen LFP signal throughout the validation period. The
agreement is maintained across the interval rather than being restricted to
samples immediately adjacent to the training boundary, indicating that the
learned latent transition operator provides accurate local predictions for
neural states occurring throughout the held-out recording.

\begin{figure*}[!t]
    \centering
    \includegraphics[width=\textwidth]{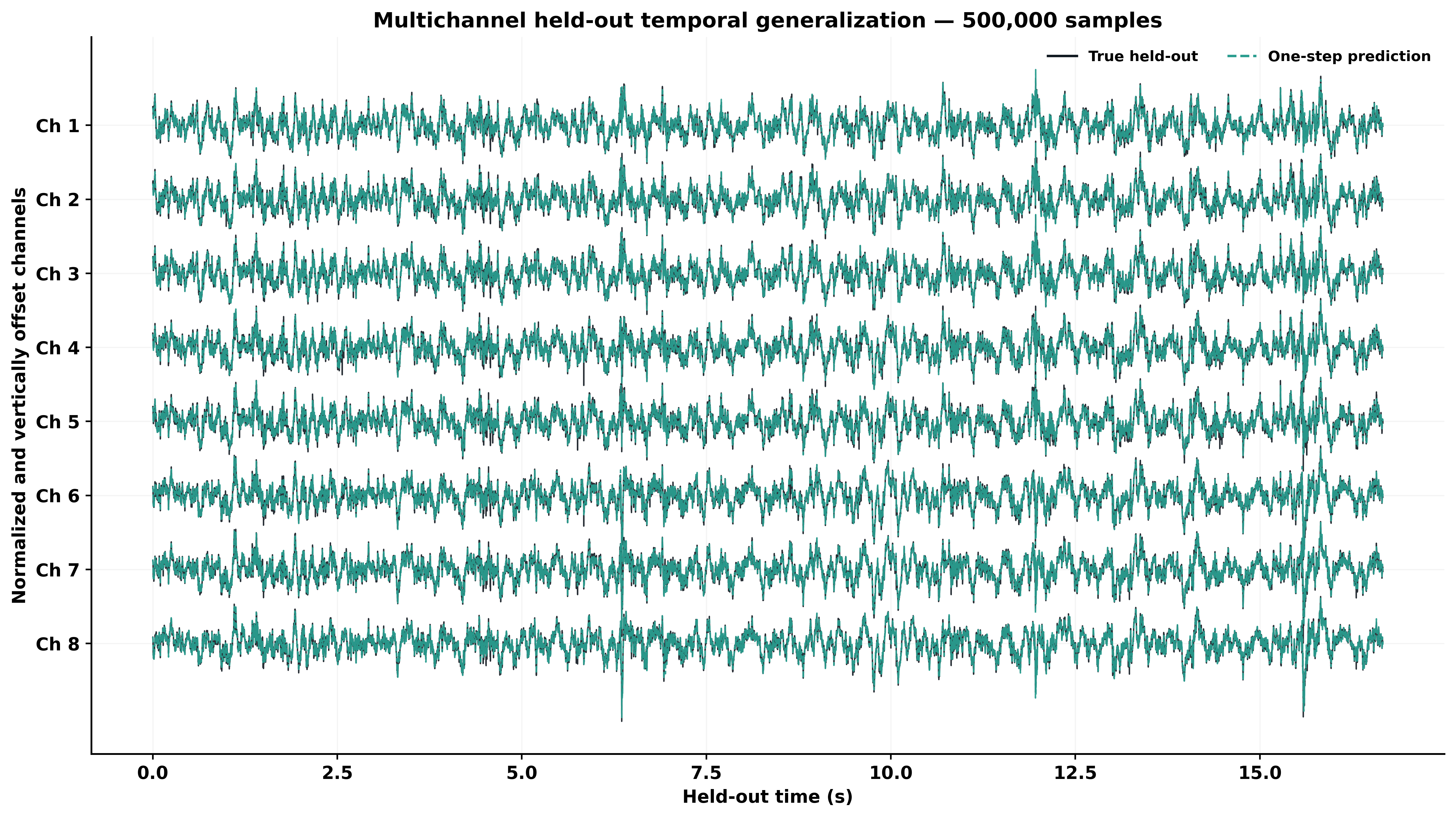}
    \caption{
    Multichannel prediction. True held-out LFP
    activity and PCA-DMD one-step predictions are shown across all channels
    over the unseen validation interval. Signals are normalized and
    vertically offset for visualization. The agreement across channels
    demonstrates accurate prediction of multivariate neural activity on
    data excluded from model fitting.
    }
    \label{fig:temporal_prediction_multichannel}
\end{figure*}

To determine whether this predictive behavior is preserved across the
spatially distributed neural recording rather than being specific to a
single channel, we further evaluated the model simultaneously across all
LFP channels. Figure~\ref{fig:temporal_prediction_multichannel} presents
the vertically offset multichannel signals over the complete held-out
interval. The predicted trajectories consistently track the corresponding
unseen signals across channels, including variations in local amplitude and
temporal structure. This multichannel agreement indicates that the prediction performance extends across the high-dimensional
neural state rather than being driven by an isolated channel.

\begin{figure*}[!t]
    \centering
    \includegraphics[width=\textwidth]{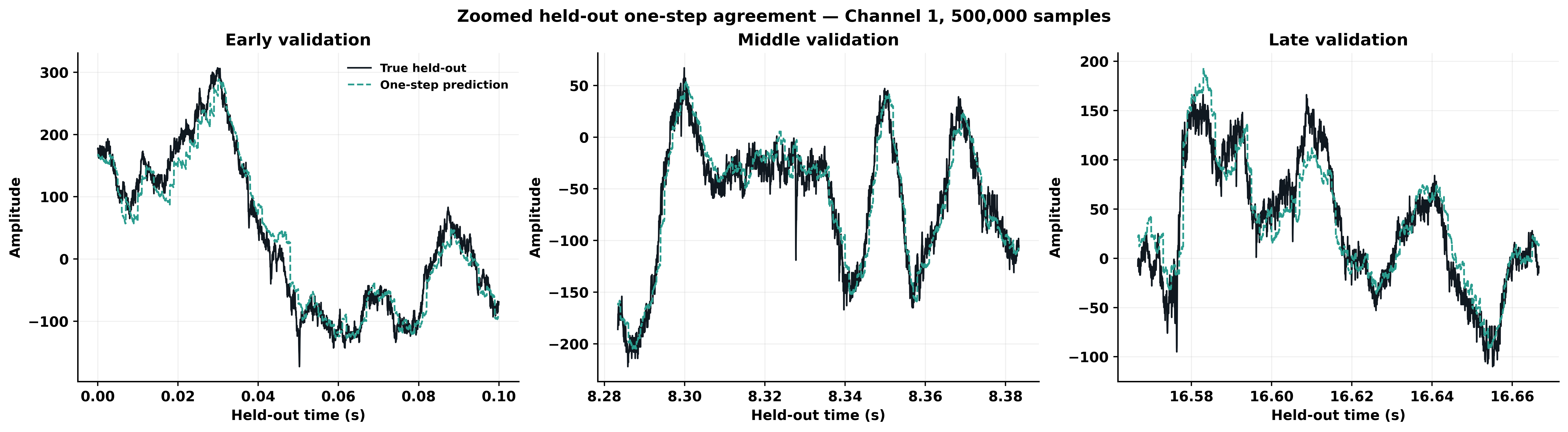}
    \caption{
    Zoomed assessment of one-step temporal prediction at early,
    middle, and late locations of the unseen validation interval. The
    PCA-DMD prediction follows the local morphology of the true LFP signal
    across temporally separated regions, providing a fine-scale view of
    prediction accuracy throughout the held-out recording.
    }
    \label{fig:temporal_prediction_zoom}
\end{figure*}

Finally, we examined short temporal regions selected from the beginning,
middle, and end of the held-out interval to provide a higher-resolution
assessment of local prediction accuracy. As shown in
Figure~\ref{fig:temporal_prediction_zoom}, the predicted trajectories
reproduce the dominant waveform morphology across all three temporal
locations. Comparable agreement is also observed in the late validation
region, which is temporally most distant from the training boundary. The
zoomed analysis therefore complements the full-interval results by showing
that the observed predictive agreement is not merely a consequence of
visual compression over long recordings, but persists at the local waveform
level throughout the unseen interval.

Together, these results demonstrate that the latent transition structure
learned by PCA-DMD supports accurate one-step prediction on temporally
held-out neural activity excluded from model fitting. The consistency
observed over the complete held-out interval, across multiple channels, and
at temporally separated local regions provides complementary evidence for
prediction of high-dimensional neural dynamics.
This experiment should be distinguished from autonomous long-horizon
forecasting: the objective is not to generate an extended future LFP
trajectory recursively from a single initial condition, but to evaluate
whether a fixed latent evolution operator learned from the training segment
can accurately predict the next state from neural states encountered in a
temporally unseen portion of the recording.

\subsection{External Validation on Allen Neuropixels LFP Recordings}
\label{sec:results_allen}

To evaluate whether the PCA-DMD reconstruction framework remains effective
beyond the primary four-subject dataset, we applied the same formulation to
an independent Allen Neuropixels LFP recording. The external recording
contained 93 channels and $300{,}000$ samples acquired at approximately
$1{,}250$ Hz, corresponding to approximately $240$ seconds of neural
activity. This experiment does not evaluate cross-dataset zero-shot
generalization; rather, it assesses whether the same PCA-DMD reconstruction
mechanism remains effective under a different channel dimensionality,
sampling rate, recording duration, and acquisition setting.

\begin{figure*}[t]
    \centering
    \includegraphics[width=.84\textwidth]{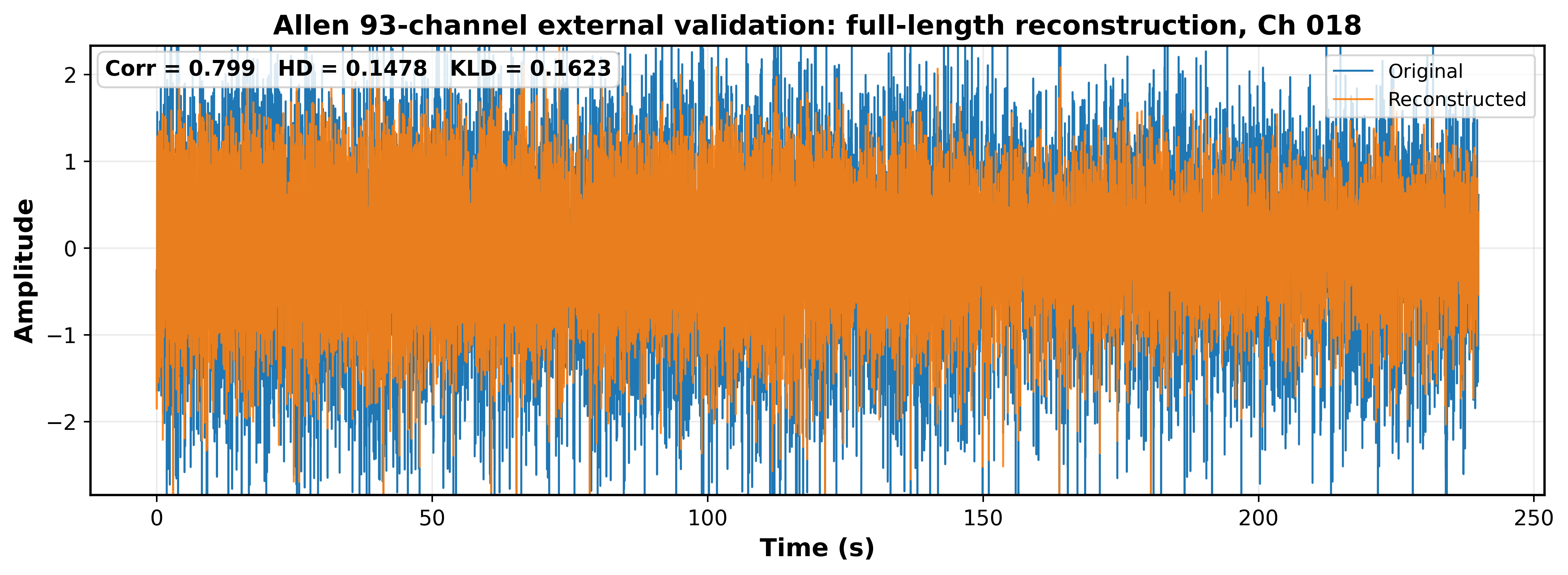}
    \caption{
    External validation on the independent 93-channel Allen Neuropixels LFP
    recording. A representative full-length reconstruction for Channel 18
    is shown together with 1000- and 2000-sample zoomed overlays and
    multichannel original-versus-reconstructed heatmaps. The representative
    channel achieved Corr$=0.799$, HD$=0.1478$, and KLD$=0.1623$.
    }
    \label{fig:results_allen_reconstruction}
\end{figure*}

Figure~\ref{fig:results_allen_reconstruction} shows a representative
full-length reconstruction for Channel 18. PCA-DMD achieved a correlation
of $0.799$, an HD of $0.1478$, and a KLD of $0.1623$ on this channel.
Despite the substantially higher channel dimensionality and different
acquisition characteristics of the external recording, the reconstruction
closely follows the amplitude envelope and dense oscillatory structure over
the complete recording.

\begin{figure*}[t]
    \centering
    \includegraphics[width=.84\textwidth]{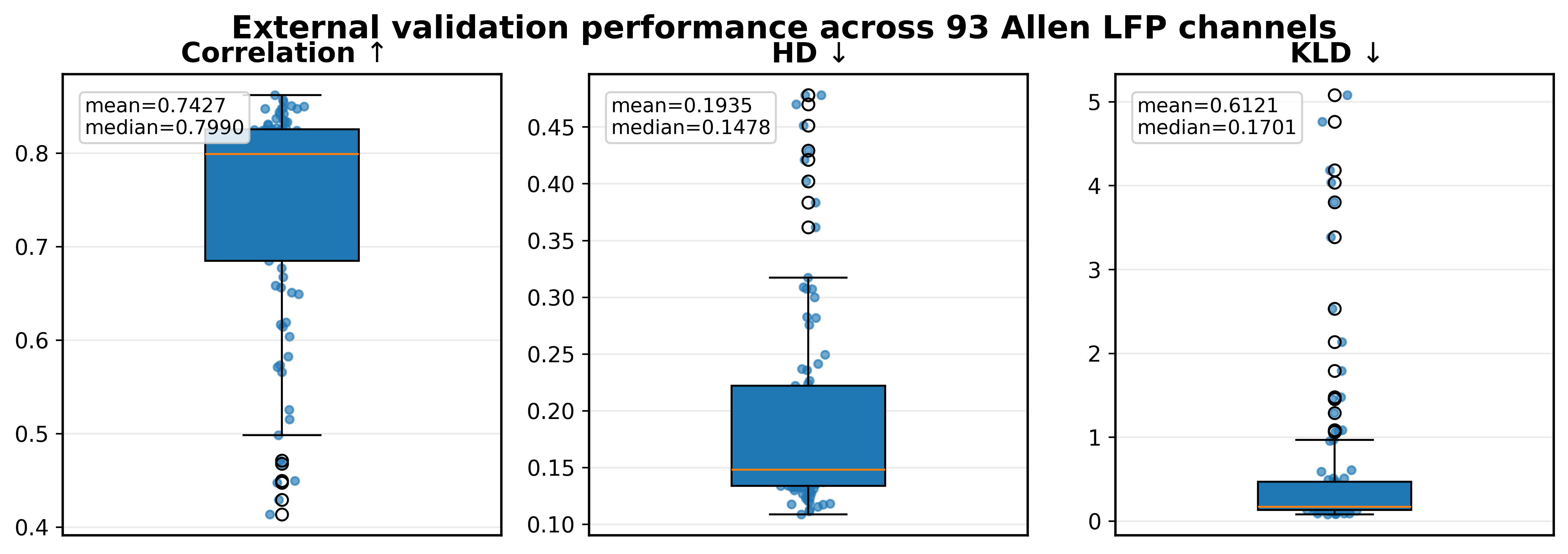}
    \caption{
    Distribution of reconstruction performance across the 93 Allen
    Neuropixels LFP channels. Boxplots and channel-level points are shown
    for correlation, HD, and KLD. The distributions show strong median
    reconstruction performance together with a subset of more challenging
    channels.
    }
    \label{fig:results_allen_metrics}
\end{figure*}

Performance across all 93 channels is summarized in
Figure~\ref{fig:results_allen_metrics}. The mean channel-wise correlation
was $0.7427$, while the median correlation was $0.7990$. The difference
between the mean and median indicates that reconstruction performance was
strong for the majority of channels but reduced for a subset of more
challenging channels. Similarly, the median HD was $0.1478$, compared with
a mean HD of $0.1935$.

The median KLD was $0.1701$, whereas the mean KLD was $0.6121$. The larger
mean reflects a relatively small number of high-divergence channels, as
indicated by the long upper tail of the channel-wise KLD distribution.
Thus, the median provides a more representative measure of reconstruction
quality for a typical Allen channel, while the mean captures the remaining
channel-dependent variability.

\subsection{Koopman Spectral and Mode-Level Analysis}
\label{sec:results_koopman}

\begin{figure*}[t]
    \centering
    \includegraphics[width=.84\textwidth]{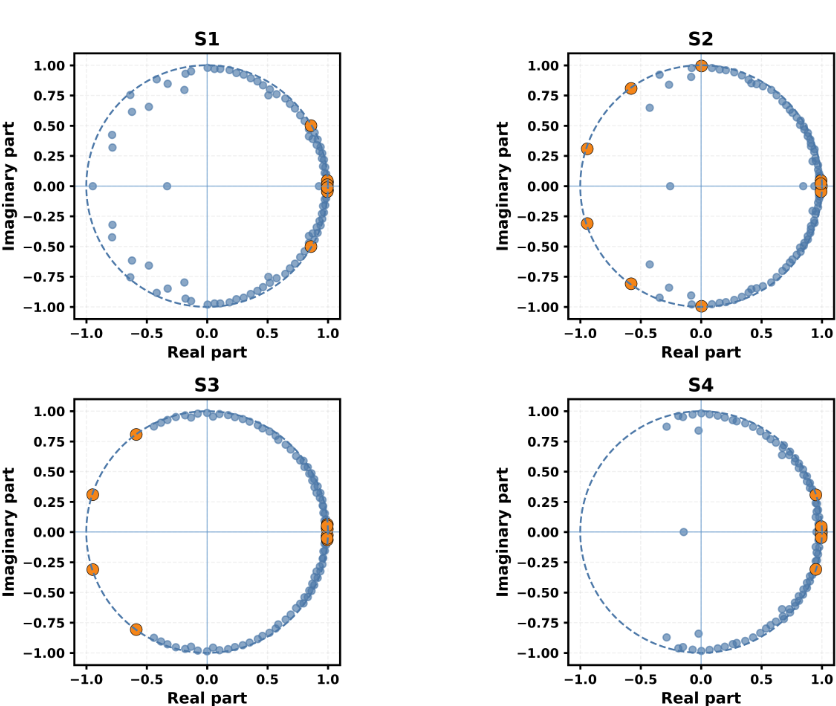}
    \caption{
    Koopman eigenvalue spectra learned by PCA-DMD across the primary
    subjects. Eigenvalues are shown in the complex plane relative to
    the unit circle. Blue markers denote the full spectrum, while
    highlighted markers indicate the dominant eigenvalues with the
    largest magnitudes. The concentration of dominant eigenvalues near
    the unit circle indicates persistent or slowly decaying latent
    dynamical components.
    }
    \label{fig:results_koopman}
\end{figure*}

We finally examined the spectral and mode-level structure of the latent
Koopman operators learned by PCA-DMD. Figure~\ref{fig:results_koopman}
shows the eigenvalue spectra across the primary subjects. The dominant
eigenvalues are concentrated near the unit circle, indicating that the
learned latent dynamics are primarily characterized by persistent or
slowly decaying components rather than strongly unstable modes. Similar
spectral organization is observed across subjects, providing a
dynamical perspective complementary to the cross-subject zero-shot
generalization results.

\begin{figure}[t]
    \centering
    \includegraphics[width=.95\columnwidth]{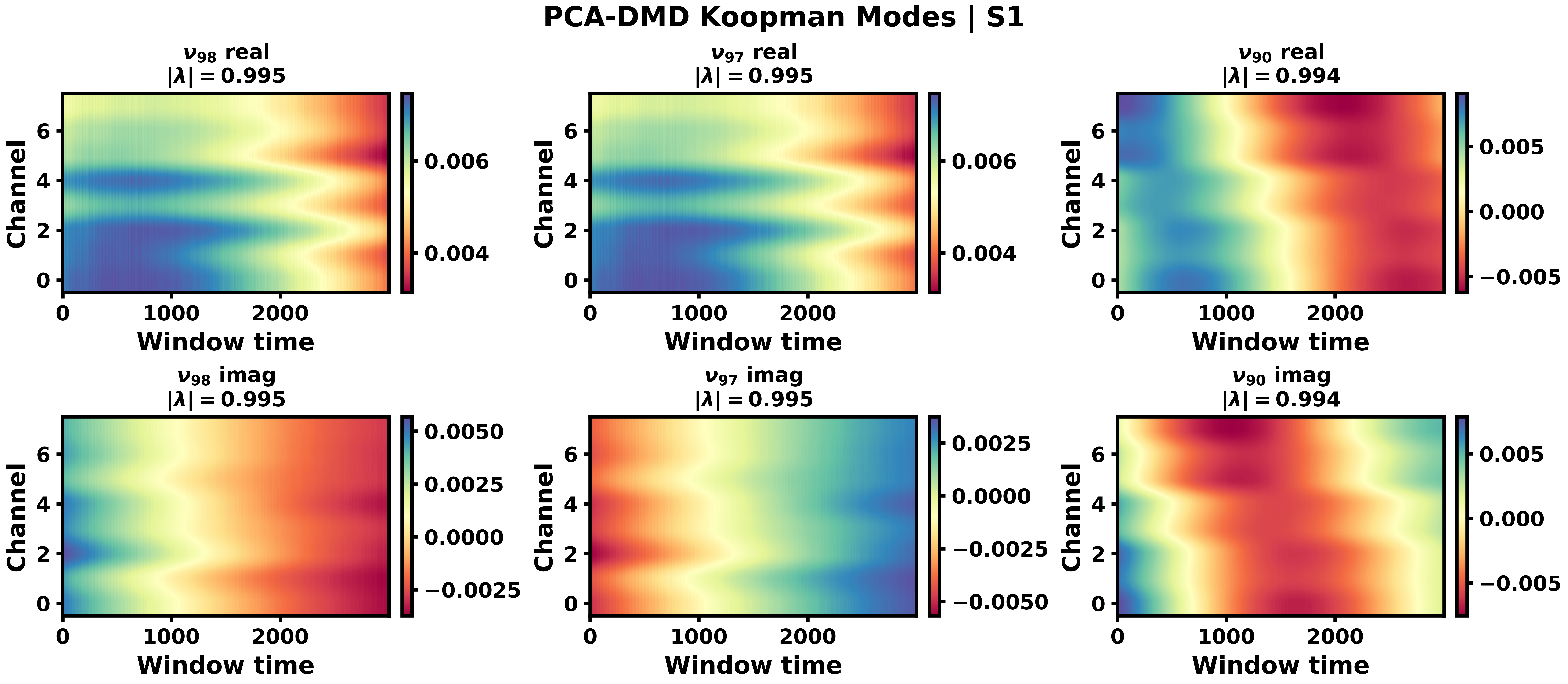}
    \caption{
    PCA-DMD Koopman mode structure for subject S1. Real and imaginary
    components of representative dominant modes are shown across window
    time and LFP channels.
    }
    \label{fig:pcadmd_s1_modes}
\end{figure}

To further characterize the structure associated with these spectral
components, we examined representative dominant Koopman modes.
Figure~\ref{fig:pcadmd_s1_modes} shows the real and imaginary components
of selected dominant modes for subject S1 across window time and the
eight LFP channels. The modes exhibit structured variation across both
temporal and channel dimensions, revealing spatiotemporal organization
within the learned latent dynamics that is not directly captured by
reconstruction metrics alone. The corresponding eigenvalue magnitudes
are close to unity, consistent with the persistent dynamical components
observed in the spectral analysis.

The external Allen Neuropixels experiment also yielded a
low-dimensional Koopman spectrum with dominant eigenvalues concentrated
near the unit circle. Thus, the spectral characterization remains
applicable when PCA-DMD is used on an independent neural recording with
substantially higher channel dimensionality. Taken together, the
eigenvalue spectra and Koopman modes provide an interpretable
dynamical characterization of the latent representation, complementing
the reconstruction, cross-subject generalization, temporal prediction,
and scalability analyses.

\section{Limitations and Future Work}
\label{sec:limitations_future}

Despite the strong reconstruction, cross-subject generalization,
prediction, and scalability results, several
limitations remain. First, PCA-DMD uses linear dimensionality reduction
and a finite-dimensional linear operator in the latent space. Although
this provides a compact and interpretable representation, strongly
nonlinear or highly localized neural events may require richer
observables than those captured by a PCA basis. Reconstruction
performance also depends on window length, step size, latent dimension,
and numerical settings, whose optimal values may vary across recording
configurations and temporal scales.

Second, the primary cross-subject experiments involve four subjects.
While the all-pair $300{,}000$-sample evaluation demonstrates strong
zero-shot generalization across these recordings, larger and more
heterogeneous cohorts are needed to establish population-level
generalization. Similarly, the scalability analysis uses S1 as the
fixed source and should ultimately be extended to multiple source
subjects, sessions, laboratories, behavioral states, and species. The
Allen Neuropixels experiment provides external validation under
substantially different acquisition characteristics, but the model is
fitted directly to the Allen recording; it therefore demonstrates
external reconstruction robustness rather than cross-dataset zero-shot
generalization.

Third, the current evaluation primarily measures signal-level
reconstruction and prediction using correlation, distributional
distances, reconstruction errors, and runtime. These metrics do not
fully establish preservation of biologically important transient
events. Event-level analyses of SWR timing, duration, ripple-band power,
phase relationships, and cross-channel propagation remain important
future validation steps. Moreover, although scalability is demonstrated
up to $900{,}000$ samples, this remains below the approximately
$22$ million samples available in the complete recordings.

The prediction experiment is also limited to one-step
prediction from observed states within a temporally held-out interval.
Because predictions are not recursively propagated from previous model
outputs, the present results should not be interpreted as evidence of
autonomous long-horizon forecasting. Evaluating multi-step and recursive
prediction, including the accumulation of error over increasing
prediction horizons, remains an important direction for future work.

Future work will extend PCA-DMD along three main directions. First,
richer latent representations based on kernel methods, nonlinear
manifold learning, autoencoders, or Koopman autoencoders could improve
the representation of nonlinear dynamics while retaining structured
latent evolution. Adaptive or multi-resolution windowing could further
accommodate neural processes occurring at different temporal scales.

Second, streaming and incremental implementations will be investigated
for full-recording analysis. Incremental PCA, online operator estimation,
randomized linear algebra, GPU acceleration, and distributed
reconstruction could enable efficient processing of complete recordings
and support near-real-time applications. Evaluation on larger
multi-subject and multi-institutional datasets will also be important
for testing cross-session, cross-probe, cross-laboratory, and direct
cross-dataset generalization.

Finally, future studies should connect reconstruction fidelity more
directly to downstream neuroscience tasks, including SWR detection and
characterization, neural-state classification, and analysis of
channel-wise propagation patterns. Incorporating uncertainty
quantification through bootstrap analysis, spectral confidence
intervals, or ensemble Koopman models could additionally identify
regions and dynamical components for which reconstruction or prediction
is less reliable.

\section{Conclusion}
\label{sec:conclusion}

We presented PCA-DMD, an operator-theoretic framework for scalable
reconstruction of long-duration multichannel neural dynamics. The method
combines overlapping temporal windows, PCA-based latent compression,
linear Koopman evolution, inverse projection, and overlap-add
aggregation to reconstruct continuous neural recordings while retaining
an interpretable spectral representation.

The experimental evaluation progressed through complementary and
increasingly demanding settings. On $200{,}000$-sample hippocampal LFP
recordings, PCA-DMD provided substantially better distributional
agreement than the evaluated conventional DMD variants, achieving a KLD
of $0.0761$ and an HD of $0.0847$. At $300{,}000$ samples, the learned
latent dynamics demonstrated strong cross-subject zero-shot
generalization across all ordered source--target subject pairs without
target-specific fine-tuning, producing correlations between $0.9504$
and $0.9800$ together with low HD and KLD values. The prediction further showed that a fixed PCA-DMD operator learned
exclusively from the training segment could accurately predict the next
state from neural states encountered in a temporally held-out interval.
Close one-step prediction agreement was maintained across the complete
unseen interval, multiple channels, and temporally separated early,
middle, and late regions.

The dedicated scalability analysis further showed that S1-based
zero-shot evaluation on S2, S3, and S4 remained stable as the signal
length increased from $400{,}000$ to $900{,}000$ samples. Mean
correlation remained approximately $0.965$--$0.968$, while HD and KLD
decreased across the tested signal lengths and runtime increased
predictably. External validation on an independent 93-channel Allen
Neuropixels recording yielded a mean channel-wise correlation of
$0.7427$ and a median correlation of $0.7990$, demonstrating that the
PCA-DMD reconstruction mechanism remains effective under a substantially
different recording configuration.

The Koopman spectra and modes complemented the reconstruction metrics by
revealing dominant eigenvalues concentrated near the unit circle and
structured temporal and channel-wise patterns in representative Koopman
modes. These analyses provide an interpretable dynamical characterization
of the learned latent representation beyond signal-level reconstruction
metrics.

Together, the results show that PCA compression provides more than
computational dimensionality reduction: it creates a compact and
well-conditioned latent space in which Koopman evolution can be
estimated stably, applied to temporally unseen states for one-step
prediction, generalized across unseen subjects without fine-tuning, and
interpreted through spectral and mode-level analysis.

Overall, PCA-DMD provides a scalable, generalizable, and interpretable
framework for reconstructing high-dimensional neural time series. By
combining within-subject reconstruction, cross-subject zero-shot
generalization, prediction, systematic
scalability analysis, independent external validation, and Koopman
spectral analysis within a unified framework, this study provides a
foundation for future operator-theoretic modeling of large-scale neural
recordings and downstream analysis of transient neural dynamics.


\bibliography{main}
\bibliographystyle{tmlr}

\newpage
\appendix
\section*{Appendix}
\section{Koopman Operator}\label{app:koop}
The Koopman operator provides a linear perspective on nonlinear dynamical systems by analyzing the evolution of \emph{observables} (functions of the state) rather than the states themselves. 
Consider a discrete-time system
\begin{align}
    \vh_{t+1} = f(\vh_t), \quad \vh_t \in \mathcal{H} \subseteq \mathbb{R}^{d_h}.
\end{align}
For an observable $\phi : \mathcal{H} \to \mathbb{C}$, the Koopman operator $\mathcal{K}$ is defined as
\begin{align*}
    [\mathcal{K}\phi](\vh) = \phi(f(\vh)).
\end{align*}
Although $f$ may be nonlinear, $\mathcal{K}$ is always linear (but typically infinite-dimensional), allowing spectral methods to be applied to nonlinear dynamics. 
Koopman eigenfunctions $\varphi_k$ and eigenvalues $\lambda_k$ satisfy
\begin{align*}
    \varphi_k(\vh_{t+1}) = \lambda_k \, \varphi_k(\vh_t),
\end{align*}
and under suitable assumptions the state can be expanded in terms of these eigenfunctions. 
This leads to the \emph{Koopman mode decomposition (KMD)}:
\begin{align}
    \vh_t \, = \, \sum_k \lambda_k^t \, \phi_{\lambda_k}(\vh_0) \, \vc_k^\Phi,
\end{align}
where $\vc_k^\Phi$ are the Koopman modes associated with the observable $\Phi$. 
Thus, KMD expresses nonlinear dynamics as a superposition of modes evolving linearly in time, forming the theoretical foundation for data-driven methods such as dynamic mode decomposition (DMD).



\section{SOTA Methods and their Results on LFP Reconstruction}
\subsection{Classical DMD}
The Classical DMD implementation mirrors the windowing approach, utilizing PyDMD's DMD with an SVD rank of 8 on the transposed windows. Reconstruction involves extracting the real part of the DMD-reconstructed data for predictions, deriving a diagonal Koopman matrix from eigenvalues, and reconstructing the full signal through overlapping window averaging tailored to the signal length.
The Classical DMD method for LFP signal reconstruction begins by windowing the signal \( x(t) \in \mathbb{R}^{N} \) into overlapping snapshots \( \mathbf{X}_{\text{full}} = \{ x(t_i : t_i + w) \}_{i=1}^{M} \), with window size \( w = 3000 \) and step \( \delta = 30 \). The transposed snapshot matrix \( \mathbf{X}_{\text{full}}^T \in \mathbb{R}^{w \times M} \) is decomposed using DMD with SVD rank \( r = 8 \), solving:
\[
\mathbf{X}_{\text{next}}^T = \mathbf{A} \mathbf{X}^T,
\]
where \( \mathbf{X} = \mathbf{X}_{\text{full}}[:-1] \), \( \mathbf{X}_{\text{next}} = \mathbf{X}_{\text{full}}[1:] \), and \( \mathbf{A} \) is approximated via DMD modes \( \mathbf{\Phi} \) and eigenvalues \( \mathbf{\Lambda} \):
\[
\mathbf{X}_{\text{full}}^T \approx \mathbf{\Phi} \mathbf{\Lambda}^t \mathbf{b},
\]
with \( \mathbf{b} \) as the initial amplitude. The predicted snapshots are \( \mathbf{X}_{\text{pred}} = \text{Re}(\mathbf{\Phi} \mathbf{\Lambda} \mathbf{b}) \). The full reconstructed signal is obtained as:
\[
x_{\text{full}}(t) = \frac{1}{c(t)} \sum_{i=1}^{M-1} \mathbf{X}_{\text{pred},i}(t - i\delta), \quad t \in [i\delta, i\delta + w),
\]
where \( c(t) \) counts overlapping windows, and the signal is trimmed to \( t \in [\delta, \min(N, T - \delta)] \).
\subsection{SpDMD}
For SpDMD, the LFP reconstruction starts with windowing the signal similarly into overlapping segments, then applying PyDMD's SpDMD with an SVD rank of 8 and a sparsity parameter rho of 1e-6 directly on the transposed window matrix. The reconstructed data from SpDMD is used to predict next snapshots, taking the real part, and a diagonal Koopman matrix is formed from the eigenvalues. Full signal reconstruction employs a custom averaging function over overlapping windows, adjusted to match the expected signal length. 

SpDMD method for LFP signal reconstruction processes a single-channel signal \( x(t) \in \mathbb{R}^{N} \) by forming overlapping windows \( \mathbf{X}_{\text{full}} = \{ x(t_i : t_i + w) \}_{i=1}^{M} \), with window size \( w = 3000 \) and step \( \delta = 30 \). The transposed snapshot matrix \( \mathbf{X}_{\text{full}}^T \in \mathbb{R}^{w \times M} \) is decomposed using SpDMD with SVD rank \( r = 8 \) and sparsity parameter \( \rho = 10^{-6} \), computing modes and eigenvalues via:
\[
\mathbf{X}_{\text{full}}^T \approx \mathbf{\Phi} \mathbf{\Lambda} \mathbf{\Phi}^{-1},
\]
where \( \mathbf{\Phi} \) contains SpDMD modes, and \( \mathbf{\Lambda} \) is a diagonal matrix of eigenvalues. The predicted snapshots are obtained as \( \mathbf{X}_{\text{pred}} = \text{Re}(\mathbf{\Phi} \mathbf{\Lambda} \mathbf{b}) \), with \( \mathbf{b} \) derived from the initial snapshot projection. Using \( \mathbf{X} = \mathbf{X}_{\text{full}}[:-1] \) and \( \mathbf{X}_{\text{next}} = \mathbf{X}_{\text{full}}[1:] \), the full reconstructed signal is computed as:
\[
x_{\text{full}}(t) = \frac{1}{c(t)} \sum_{i=1}^{M-1} \mathbf{X}_{\text{pred},i}(t - i\delta), \quad t \in [i\delta, i\delta + w),
\]
where \( c(t) \) is the count of overlapping windows at time \( t \), and the signal is trimmed to \( t \in [\delta, \min(N, T - \delta)] \).
 
\subsection{HODMD}
HODMD for LFP reconstruction incorporates data scaling with StandardScaler before fitting PyDMD's HODMD with SVD rank 8 and delay d=2 on scaled transposed windows. The real part of reconstructed data is inverse-scaled, and predictions are made on shifted windows. Signal rebuilding uses the same overlapping averaging method. 
HODMD method for LFP reconstruction begins by windowing the signal \( x(t) \in \mathbb{R}^{N} \) into overlapping snapshots \( \mathbf{X}_{\text{full}} = \{ x(t_i : t_i + w) \}_{i=1}^{M} \), with window size \( w = 3000 \) and step \( \delta = 30 \), scaled to \( \tilde{\mathbf{X}}_{\text{full}} = \text{StandardScaler}(\mathbf{X}_{\text{full}}) \). HODMD, with SVD rank \( r = 8 \) and delay \( d = 2 \), is applied to the transposed snapshot matrix \( \tilde{\mathbf{X}}_{\text{full}}^T \in \mathbb{R}^{w \times M} \), constructing an augmented Hankel matrix and decomposing it as:
\[
\tilde{\mathbf{X}}_{\text{full}}^T \approx \mathbf{\Phi} \mathbf{\Lambda}^t \mathbf{b},
\]
where \( \mathbf{\Phi} \) and \( \mathbf{\Lambda} \) are the DMD modes and eigenvalues, and \( \mathbf{b} \) is the amplitude vector. The reconstructed snapshots \( \tilde{\mathbf{X}}_{\text{rec}} \) are inverse-scaled to \( \mathbf{X}_{\text{rec}} \). Predictions use \( \mathbf{X} = \mathbf{X}_{\text{full}}[:-1] \), with \( \mathbf{X}_{\text{pred}} = \text{Re}(\text{HODMD}(\text{StandardScaler}(\mathbf{X}))^T) \). The full signal is reconstructed via:
\[
x_{\text{full}}(t) = \frac{1}{c(t)} \sum_{i=1}^{M-1} \mathbf{X}_{\text{pred},i}(t - i\delta), \quad t \in [i\delta, i\delta + w),
\]
where \( c(t) \) counts overlapping windows, and the signal is trimmed to \( t \in [\delta, \min(N, T - \delta)] \).

\subsection{MrDMD}
MrDMD implementation scales the windowed data and employs PyDMD's MrDMD with a base DMD of SVD rank 8, max level=2, and max cycles=1, fitting on scaled transposed matrices for multi-scale decomposition. Reconstructions and predictions use the real part inverse-scaled, with full signal assembly via overlapping averages. 
MrDMD method processes the LFP signal \( x(t) \in \mathbb{R}^{N} \) by forming overlapping windows \( \mathbf{X}_{\text{full}} = \{ x(t_i : t_i + w) \}_{i=1}^{M} \), with window size \( w = 3000 \) and step \( \delta = 30 \), scaled to \( \tilde{\mathbf{X}}_{\text{full}} = \text{StandardScaler}(\mathbf{X}_{\text{full}}) \). The MrDMD model, with a base DMD of SVD rank \( r = 8 \), maximum decomposition level \( L = 2 \), and maximum cycles \( C = 1 \), decomposes the transposed snapshot matrix \( \tilde{\mathbf{X}}_{\text{full}}^T \in \mathbb{R}^{w \times M} \) into multi-scale modes:
\[
\tilde{\mathbf{X}}_{\text{full}}^T \approx \sum_{l=1}^{L} \sum_{c=1}^{C} \mathbf{\Phi}_{l,c} \mathbf{\Lambda}_{l,c}^t \mathbf{b}_{l,c},
\]
where \( \mathbf{\Phi}_{l,c} \), \( \mathbf{\Lambda}_{l,c} \), and \( \mathbf{b}_{l,c} \) are the modes, eigenvalues, and amplitudes at level \( l \) and cycle \( c \). The reconstructed snapshots \( \tilde{\mathbf{X}}_{\text{rec}} \) are inverse-scaled to \( \mathbf{X}_{\text{rec}} \), and predictions use \( \mathbf{X} = \mathbf{X}_{\text{full}}[:-1] \) and \( \mathbf{X}_{\text{pred}} = \text{Re}(\text{MrDMD}(\text{StandardScaler}(\mathbf{X}))^T) \). The full signal is reconstructed as:
\[
x_{\text{full}}(t) = \frac{1}{c(t)} \sum_{i=1}^{M-1} \mathbf{X}_{\text{pred},i}(t - i\delta), \quad t \in [i\delta, i\delta + w),
\]
with \( c(t) \) as the overlap count, trimmed to \( t \in [\delta, \min(N, T - \delta)] \).

\section{(Hyper)parameters}

\begin{table}[h]
\centering
\caption{Short Forms of Parameters}
\begin{tabular}{ll}
\toprule
\textbf{Parameter} & \textbf{Short Form} \\
\midrule
Sampling Frequency & fs \\
Maximum Samples & max\_samples \\
Window Size & window\_size \\
Step Size & step \\
Latent Dimension & latent\_dim \\
Sparsity Parameter & rho \\
Delay Embeddings & d \\
Maximum Decomposition Levels & max\_level \\
Maximum Cycles & max\_cycles \\
\bottomrule
\end{tabular}
\end{table}
\begin{table}[h]
\centering
\caption{Parameters Used in PCA-DMD, Classical DMD, SpDMD, HODMD, and MrDMD}
\begin{tabular}{lccccc}
\toprule
\textbf{Parameter} & \textbf{PCA-DMD} & \textbf{Classical DMD} & \textbf{SpDMD} & \textbf{HODMD} & \textbf{MrDMD} \\
\midrule
\texttt{fs} & 30,000 & 30,000 & 30,000 & 30,000 & 30,000 \\
\texttt{max\_samples} & 200,000 & 200,000 & 200,000 & 200,000 & 200,000 \\
\texttt{window\_size} & 3,000 & 3,000 & 3,000 & 3,000 & 3,000 \\
\texttt{step} & 30 & 30 & 30 & 30 & 30 \\
\texttt{latent\_dim} & 8 & 8 & 8 & 8 & 8 \\
\texttt{rho} & - & - & 1e-8 & - & - \\
\texttt{d} & - & - & - & 50 & - \\
\texttt{max\_level} & - & - & - & - & 5 \\
\texttt{max\_cycles} & - & - & - & - & 2 \\
\bottomrule
\end{tabular}
\end{table}

\section{Evaluation Measures}\label{app:mea}
\subsection{Kullback-Leibler divergence (KLD)}
To evaluate the geometrical agreement between the true and reconstructed LFP signals in  PCA-DMD and DMD-based methods (Classical DMD, SpDMD, MrDMD, and HODMD), we employed the KLD as a state space divergence metric. Specifically, for each of the eight LFP channels, we estimated probability distributions \( p(x) \) and \( q(x) \) from the trajectories of the true and reconstructed signals, respectively, in the observation space. The KLD was computed using histograms with 100 bins over the trimmed signal segments to approximate the distributions, with a small regularization term (\( \epsilon = 10^{-10} \)) added to avoid numerical issues. The KLD, averaged across channels, quantifies the discrepancy between the true and reconstructed attractor geometries, with lower values indicating better fidelity. Mathematically, the state space divergence is defined as:
\[
KLD_{\text{}}(p(x) \parallel q(x)) = \int_{\mathbb{R}^N} p(x) \log \frac{p(x)}{q(x)} \, dx,
\]
where \( p(x) \) and \( q(x) \) represent the probability densities functions of the true and reconstructed trajectories, respectively.
\subsection{Hellinger Distance}\label{app:H}
To assess the temporal agreement between the ground truth and reconstructed LFP signals, we utilized the (\( HD \)) as a temporal measure, bounded between $0$ and $1$,
averaged across all eight dynamical variables (channels). For each channel, we computed the power spectra \( f_i(\omega) \) and \( g_i(\omega) \) for the true and reconstructed signals, respectively, using histogram-based approximations with 100 bins over the trimmed signal segments, normalized to satisfy \( \int_{-\infty}^{\infty} f_i(\omega) \, d\omega = 1 \) and \( \int_{-\infty}^{\infty} g_i(\omega) \, d\omega = 1 \). A regularization term (\( \epsilon = 10^{-10} \)) was applied to ensure numerical stability. The Hellinger distance, ranging from 0 (perfect agreement) to 1, was calculated per channel and averaged to produce HD.
The Hellinger distance for the \( i \)-th channel is defined as:
\[
HD(f_i(\omega), g_i(\omega)) = \sqrt{1 - \int_{-\infty}^{\infty} \sqrt{f_i(\omega) g_i(\omega)} \, d\omega},
\]
where \( f_i(\omega) \) and \( g_i(\omega) \) are the normalized power spectra of the true and reconstructed signals for the \( i \)-th channel.





\end{document}